\documentclass[11pt]{article}
\pdfoutput=1
\usepackage{acl}
\usepackage{times}
\usepackage{latexsym}
\usepackage[T1]{fontenc}
\usepackage[utf8]{inputenc}
\usepackage{xcolor}
\usepackage{graphicx} 
\usepackage{booktabs}
\usepackage{authblk}
\usepackage{multicol}
\usepackage{multirow}
\usepackage{subcaption}

\newcommand{\ra}[1]{\renewcommand{\arraystretch}{#1}}

\title{A Multi-Aspect Framework for Counter Narrative Evaluation using Large Language Models\\ \vspace{0.5em}
\author[]{Jaylen Jones}
\author[]{Lingbo Mo}
\author[]{Eric Fosler-Lussier}
\author[]{Huan Sun}
\affil[]{The Ohio State University \\
\{jones.6278, mo.169, sun.397\}@osu.edu; fosler@cse.ohio-state.edu}
{\large \textcolor{red}{Content Warning: This paper contains potentially offensive and harmful text. }}}

\date{December 2023}

\begin{document}

\maketitle

\begin{abstract}
Counter narratives --- informed responses to hate speech contexts designed to refute hateful claims and de-escalate encounters --- have emerged as an effective hate speech intervention strategy. While previous work has proposed automatic counter narrative generation methods to aid manual interventions, the evaluation of these approaches remains underdeveloped. Previous automatic metrics for counter narrative evaluation lack alignment with human judgment as they rely on superficial reference comparisons instead of incorporating key aspects of counter narrative quality as evaluation criteria. To address prior evaluation limitations, we propose a novel evaluation framework prompting LLMs to provide scores and feedback for generated counter narrative candidates using 5 defined aspects derived from guidelines from counter narrative specialized NGOs. We found that LLM evaluators achieve strong alignment to human-annotated scores and feedback and outperform alternative metrics, indicating their potential as multi-aspect, reference-free and interpretable evaluators for counter narrative evaluation.\footnote[1]{Our code is available at \url{https://github.com/OSU-NLP-Group/LLM-CN-Eval}.}
 
\end{abstract}

\begin{figure}[t]
    \centering
    \includegraphics[width=\columnwidth]{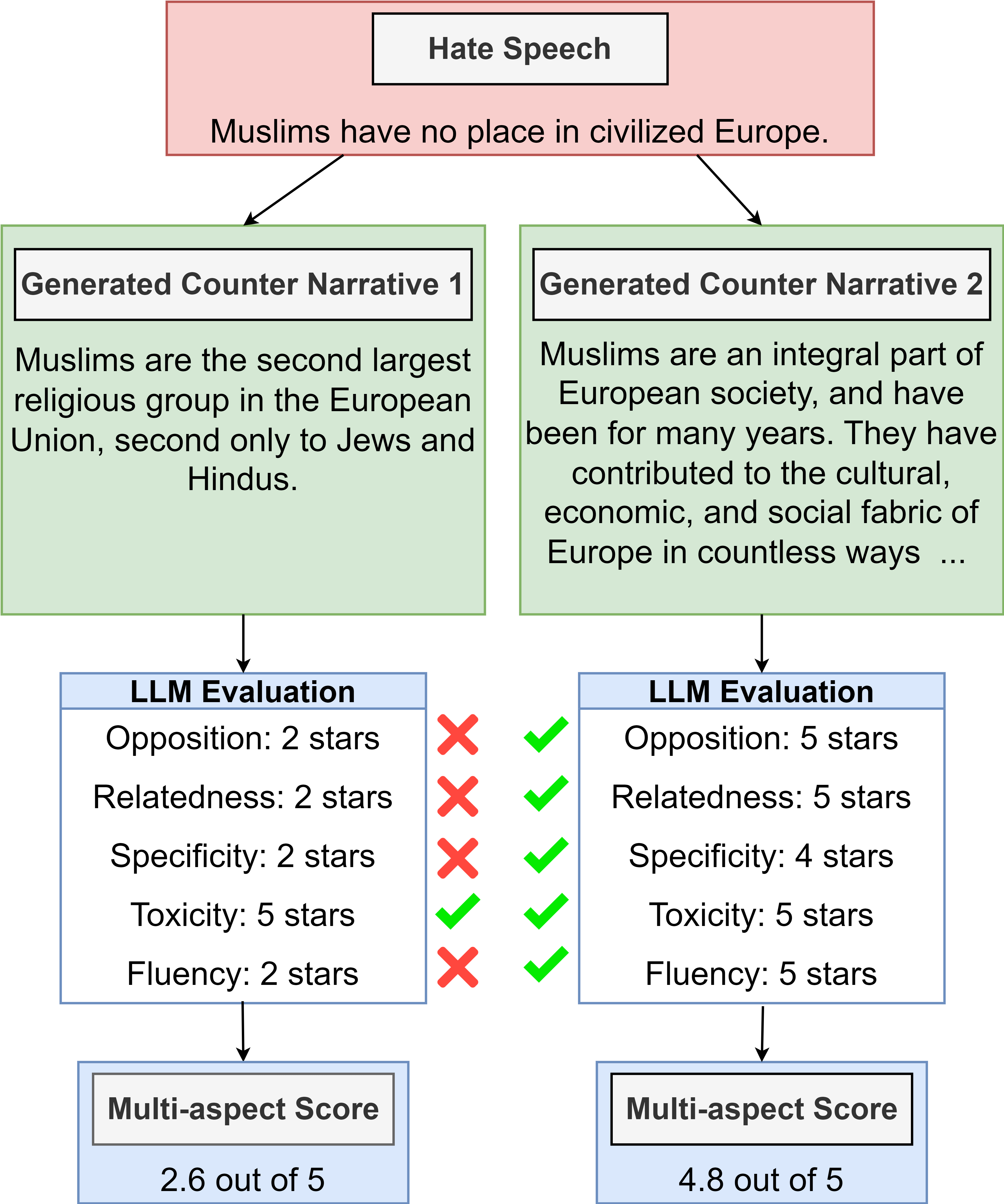}
    \caption{Example of our multi-aspect counter narrative evaluation framework.}
    \label{fig:intro}
\end{figure}

\begin{figure*}[ht]
    \centering
    \includegraphics[width=\linewidth]{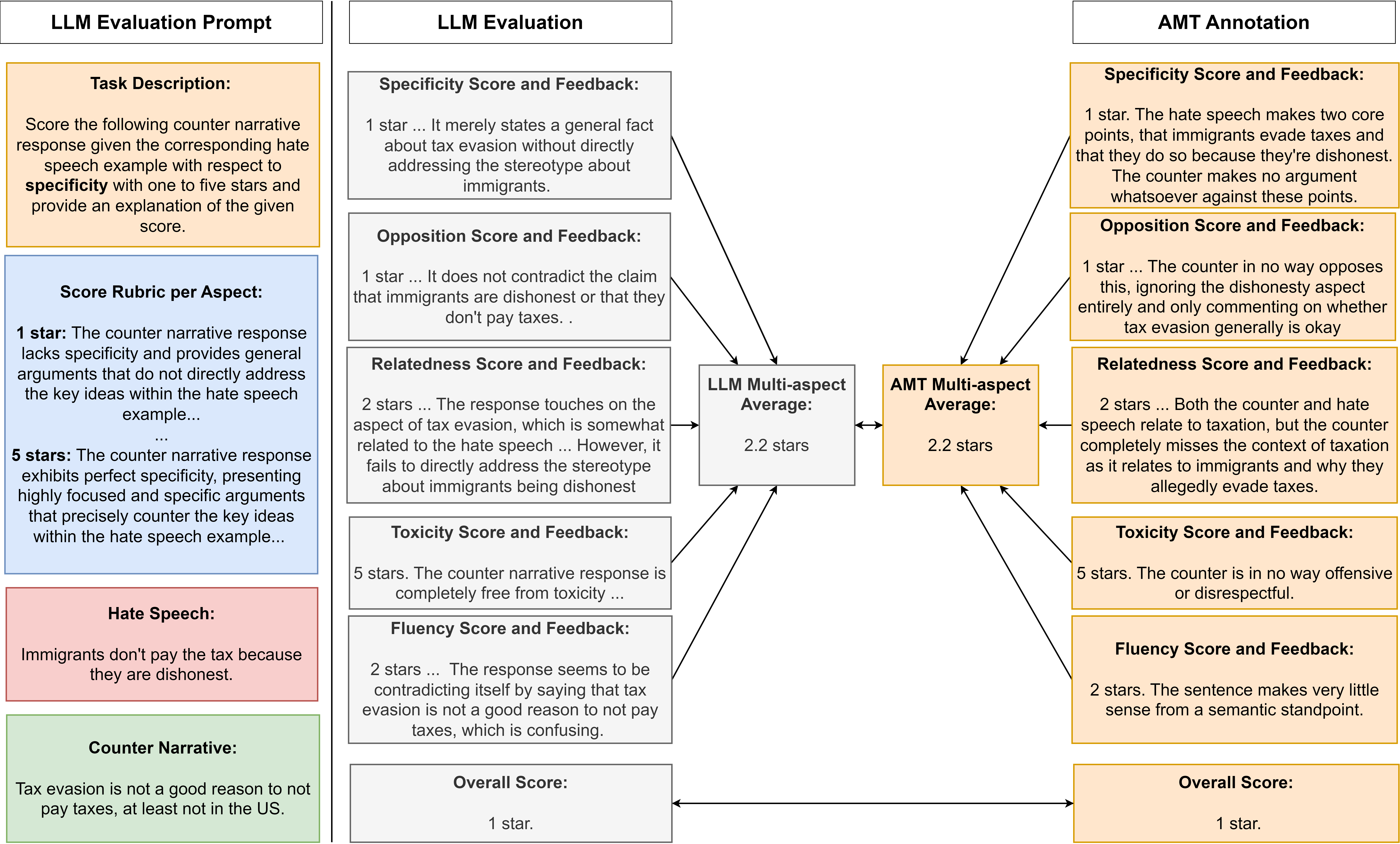}
    \caption{Validation pipeline for our counter narrative evaluation framework. (Left) Evaluation prompt template including task description, a ChatGPT-generated aspect score rubric, and hate speech/counter narrative pair. (Right) LLM evaluation scores are generated for counter narratives and are compared to AMT-annotated evaluation.}
    \label{fig:methodology}
\end{figure*}

\section{Introduction}
As online platforms allow for rapid and widespread dissemination of hate speech, automatic intervention strategies have become a growing necessity. Counter narratives --- informed responses to hate speech designed to refute hateful claims and de-escalate encounters --- have gained attention for challenging such content while minimizing free speech infringement concerns in content removal strategies. Despite the establishment of numerous NGOs\footnote[2]{\url{https://getthetrollsout.org/}} for hate speech intervention using counter narratives, effective manual intervention is impractical due to a constant influx of online toxicity. 

To augment manual intervention, numerous counter narrative generation approaches have emerged, but evaluation remains difficult. Metrics like BLEU \citep{papineni2002bleu} and ROUGE-L \citep{li2016diversity} can misalign with human judgment, as references only implicitly define the key aspects of good counter narratives. On the other hand, human evaluation using trained workers is costly and time-consuming. Previous work has used LLMs for aspect-based evaluation to address similar limitations in  tasks like summarization, but has overlooked their application in socially-oriented tasks, raising questions about their effectiveness in representing guidelines requiring social understanding \citep{magooda2023framework}.

We propose a novel multi-aspect counter narrative evaluation framework leveraging the capabilities of pretrained LLMs to determine the quality of counter narrative candidates (Figure \ref{fig:intro}).  LLMs provide evaluation scores and feedback based on five key aspects inspired by NGO guidelines:  specificity, opposition, relatedness, toxicity, and fluency. This approach improves alignment with human judgment while generating interpretable feedback and reducing reference reliance. We validate our evaluation framework by correlating LLM-generated scores with human-annotated scores and qualitatively analyzing feedback.

\section{Related Work}

Previous studies produced automatic counter narrative generation using counter narrative datasets \citep{mathew2018analyzing, qian2019benchmark, chung2019conan, bonaldi2022human} or prompting strategies \citep{ashida2022towards, zheng-etal-2023-makes}. The most prominent dataset is the expert-annotated Multitarget-CONAN\footnote[3]{\url{github.com/marcoguerini/CONAN}} \citep{fanton2021human} which contains pairs for various target groups. For evaluation, previous studies \citep{ashida2022towards, chung2021towards, zhu2021generate, bonaldi2023weigh} used overlap and diversity metrics \citep{papineni2002bleu, lin2004rouge, wang2018sentigan, li2016diversity} in addition to trained human annotation. \citet{tekirouglu2022using} uses BLEU and ROUGE-L to select a model's best generated candidate before conducting other automatic and human evaluation. However, these metrics are incapable of representing key, NGO-specified aspects of counter narrative quality solely with reference comparisons. Meanwhile, the annotator training procedure established by \citet{fanton2021human} relies directly on the use of expert NGO operators which can be costly, time-consuming, and difficult to reproduce. In other generation tasks, SoTA LLMs have been leveraged to evaluate generated candidates, leading to better alignment with human judgment \citep{chiang-lee-2023-large, liu2023gpteval, zheng2023judging, hsu2023gpt, kim2024prometheus}. \citet{jiang2023raucg} previously used GPT-3.5 for counter narrative evaluation; we develop a novel evaluation framework by prompting LLMs with NGO-derived aspects to generate scores and feedback. We validate our framework against human evaluation, enabling an interpretable and multi-aspect approach. 

\begin{table}[t]
    \centering
    \resizebox{\linewidth}{!}{%
    \begin{tabular}{ c }
         \hline 
        Evaluation Aspects \\
        \hline
        \multirow{5}{*}{\parbox{25em}{\textbf{Specificity} - how much the counter narrative presents focused and specific arguments that effectively counter the key ideas within the hate speech example through the use of in-depth arguments, nuanced reasoning, and supporting evidence.}} \\
        \\
        \\
        \\
        \\
        \multirow{5}{*}{\parbox{25em}{\textbf{Opposition} - how much the counter narrative opposes and contradicts the hate speech example through the expression of an opposing sentiment regardless of the argument’s effectiveness or persuasiveness.}} \\
        \\
        \\
        \\
        \\
        \multirow{3}{*}{\parbox{25em}{\textbf{Relatedness} - the association between the counter narrative response and hate speech example based on contextual or semantic similarity.}} \\
        \\
        \\
        \multirow{3}{*}{\parbox{25em}{\textbf{Toxicity} - how rude, unreasonable, or disrespectful a response is; definition from \citet{tekirouglu2022using};}} \\
        \\ 
        \\
        \multirow{3}{*}{\parbox{25em}{\textbf{Fluency} - the quality of a response based on whether they are well-written and grammatically correct; definition from \citet{fu2023gptscore}.}}\\
        \\
        \\
        \hline
    \end{tabular}
    }
    \caption{Key evaluation aspects used in our counter narrative evaluation framework.}
    \label{table:eval_aspects}
\end{table}

\begin{table}[t]
\ra{1.5}
\huge
\centering
\resizebox{\columnwidth}{!}{%
\begin{tabular}{@{}l l l l  l l l @{}}
 \hline
 \multicolumn{7}{c}{Evaluation Metric Correlations} \\
 \hline
 \multirow{2}{4cm}{Metric} &\multicolumn{3}{c}{AMT Multi-aspect}  &  \multicolumn{3}{c}{AMT Overall} \\
 & Pear. & Spear. & Kend. &  Pear. & Spear. & Kend. \\
\hline
BLEU1 & -0.041 & -0.102 & -0.071 & -0.048 & -0.083 & -0.06  \\
BLEU3  & 0.014  & -0.085 & -0.075 &  0.001  & -0.083 & -0.071 \\
BLEU4  & -0.032 & -0.187 & -0.141 &  -0.04  & -0.187 & -0.143 \\
ROUGE-L  & -0.052 & -0.111 & -0.079 &  -0.092 & -0.122 & -0.087 \\
BERTScore  & -0.099 & -0.092 & -0.062 & -0.102 & -0.089 & -0.063 \\
BARTScore - Recall  & 0.581  & 0.565  & 0.405  & 0.596  & 0.564  & 0.417  \\
\hline
ChatGPT Multi-Aspect& 0.664  & 0.626  & 0.481  & 0.632  & 0.609  & 0.475  \\
ChatGPT Overall & 0.658  & 0.633  & 0.517  & 0.654  & 0.624  & 0.521  \\
Vicuna-33b v.1.3 Multi-Aspect & \textbf{0.824}  & \textbf{0.782}  & \textbf{0.613}  & \textbf{0.815}  & \textbf{0.771}  & \textbf{0.616}  \\
Vicuna-33b v.1.3 Overall & 0.718  & 0.698  & 0.544  & 0.745  & 0.687  & 0.544  \\
GPT-4 Multi-Aspect & \underline{0.806}  & 0.710   & 0.557  & 0.762  & 0.694  & 0.551  \\
GPT-4 Overall & 0.788 & \underline{0.733} & \underline{0.597}  & \underline{0.783}  & \underline{0.721}  & \underline{0.600} \\
Prometheus-13b Multi-Aspect & 0.784 & 0.671 & 0.510 & 0.763 & 0.643 & 0.495 \\
Prometheus-13b Overall & 0.679 & 0.567 & 0.458 & 0.667 & 0.570 & 0.468 \\
\end{tabular}
}
\caption{Correlation of evaluation metric and AMT scores for the entire evaluation set; best correlation is in \textbf{bold}, second is \underline{underlined}. Full version in Table \ref{table:correlation_full}}
\label{table:correlation}
\end{table}

\definecolor{green}{HTML}{009B55}

\begin{table*}[t]
\centering
\resizebox{\linewidth}{!}{%
\begin{tabular}{@{}l  p{40em} @{}}
\hline 
\multicolumn{2}{c}{Aligned Counter Narrative Evaluation} \\
\hline
 \multirow{2}{4cm}{Hate Speech} & \multirow{2}{*}{\parbox{40em}{LGBTs just want to brainwash our children. They should not be taught about gender identity and sexual orientation in school.}} \\
 & \\
  \hline
 \multirow{2}{4cm}{DialoGPT Generated Counter Narrative} & \multirow{2}{*}{\parbox{40em}{People have the right to be interested in their own gender identity and sexual orientation.
}} \\
 & \\
 \hline
 \multirow{3}{4cm}{Human Specificity Score and Feedback} & \multirow{3}{*}{\parbox{40em}{2 stars. The counter entirely \textcolor{green}{ignores the claim that LGBT people want to brainwash people, and it only somewhat and indirectly touches upon the idea that schools shouldn't be teaching these topics in school} by suggesting people have a right to be interested in their own sexuality.}} \\
 & \\
 & \\
 \hline
 \multirow{3}{4cm}{Vicuna-generated Specificity Score and Feedback} & \multirow{3}{*}{\parbox{40em}{2 stars … While it acknowledges the right to explore one's gender identity and sexual orientation, \textcolor{green}{it does not directly counter the claim that LGBT individuals are trying to ``brainwash" children or that they should not be taught about these topics in school} …
}} \\
& \\
& \\
\hline
\end{tabular}
}
\caption{Example of effective counter narrative evaluation from LLM evaluator.}
\label{table:aligned_evaluation}
\end{table*}

\section{Data and Methodology}
For evaluation, we define key aspects of counter narrative quality shown in Table \ref{table:eval_aspects}. We specifically draw inspiration from NGO guidelines that advocate for constructive, focused counter narrative responses that challenge hate speech claims while de-escalating encounters in a non-toxic manner. From this, we derive specificity and relatedness, focusing on the association between the counter narrative arguments and the hate speech claims; opposition, focusing on how effectively the counter narrative denounces the hate speech; toxicity, focusing on responding civilly and positively; and fluency, focusing on the coherence of the response. By directly integrating these aspects within our LLM evaluation framework through the use of prompting, we allow for an automatic evaluation approach that is directly predicated on relevant characteristics of counter narrative quality as its criteria. 

We generate counter narratives to 180 Multitarget-CONAN test set examples using (1) DialoGPT trained on 4003 examples, the best model in \citet{tekirouglu2022using}, (2) zero-shot prompted ChatGPT \cite{chatgpt} and (3) Vicuna \citep{vicuna2023} as closed/open-source model representatives. We evaluate these generated examples with our approach and measure the correlation to human-generated scores.  While previous counter narrative work have utilized trained expert annotators for hate speech/counter narrative pair post-editing and evaluation \cite{fanton2021human}, we are unable to reproduce this process due to a lack of direct access to expert NGO operators. As an alternative, each candidate counter narrative in our study is evaluated by Amazon Mechanical Turk (AMT) workers to represent human interpretation of NGO guidelines for the task. Non-expert annotation from AMT can often be less reliable than evaluation from more trusted sources; in order to address this limitation, we conduct an extensive qualification and monitoring procedure. All workers within our study must complete a qualification task as shown in Figure \ref{fig:amt_qual_description} that involves reading training material describing what a counter narrative is, a description of the evaluation task, our evaluation aspects, and hate speech/counter narrative examples. Each worker must then pass the qualification test shown in Figures \ref{fig:amt_qual_questions} and \ref{fig:amt_qual_task} before being able to provide any evaluation. In addition, we maintained active communication with each worker throughout the study and manually verified each provided evaluation score and feedback to ensure the task is performed in an appropriate manner. As a result, we ensured high-quality annotation in our study despite the lack of expert NGO operators; additional AMT study details can be found in Appendix \ref{appendix:E}.

For automatic evaluation, we evaluate each candidate with a single run of ChatGPT, Vicuna, GPT-4 \citep{openai2023gpt4}, and Prometheus \citep{kim2024prometheus} using the evaluation prompt shown in Figure \ref{fig:methodology}. Both human and LLM evaluations result in a 1-5 star score per aspect that is aggregated into a multi-aspect average and a final 1-5 star overall score. We also collect explanations from the AMT workers and LLM evaluators to allow for a qualitative comparison of their score justifications, allowing us to verify whether LLM evaluators are right for the right reasons. We also evaluate each example using automatic metrics: BLEU, ROUGE-L, METEOR \citep{banerjee2005meteor}, BERTScore \citep{zhang2019bertscore}, and BARTScore \citep{yuan2021bartscore} using Multitarget-CONAN examples as references for comparison to alternative metrics. 

\section{Results}

\subsection{Evaluation Metric Correlation}
We measure the correlation between automatic and AMT-annotated evaluation scores using Pearson, Spearman, and Kendall coefficients to represent alignment of each evaluation metric to human judgment, presenting our results in Table \ref{table:correlation}. The overlap metrics used in previous studies achieve poor or negative correlations for our evaluation set. BERTScore's more advanced reference comparison also achieves poor correlations, suggesting that counter narrative references may not effectively represent NGO guidelines. BARTScore using Recall (described in Appendix D) achieves strong correlations; correlations for more variations are shown in Table \ref{table:correlation_full}. \textbf{LLM evaluators achieve the highest correlations with AMT-annotated evaluation scores} due to directly evaluating relevant aspects of counter narrative quality. This suggests that LLM evaluators can serve as a better alternative for counter narrative evaluation with improved alignment while offering interpretability and alleviating reference reliance. \textbf{In addition, our multi-aspect framework leads to improved evaluation performance for open-source models} and allows for Vicuna to achieve comparable performance to GPT-4. Our interpretation of multi-aspect improvement within our evaluation framework for open-source models is discussed in our qualitative evaluation (Sec. 4.3). 

\subsection{Fine-grained Analysis}
We also measure correlations per counter narrative generation model to assess robustness to generation approach and associated stylistic changes (shown in in Tables \ref{table:fine_grained_dialogpt}, \ref{table:fine_grained_chatgpt} and \ref{table:fine_grained_vicuna}). While we computed correlations for ChatGPT-generated candidates, the low variance in their AMT scores makes the correlations uninformative. \\
\noindent \textbf{DialoGPT.} Overlap based metrics are relatively more effective, indicating their viability for evaluating supervised models trained on the same distribution as the references used in evaluation as done in \citet{tekirouglu2022using}. However, LLM evaluators still outperform these metrics. Notably, ChatGPT, Vicuna, and Prometheus show decreased performance, particularly when using the overall score rather than our multi-aspect approach. \\
\noindent \textbf{Vicuna.} LLM evaluators show decreased performance for Vicuna-generated candidates, allowing BARTScore variants to achieve near-best performance. Despite achieving top performance in all other cases, GPT-4 underperforms when evaluating only Vicuna-generated candidates. The justifications for both findings are discussed in our qualitative evaluation (Sec 4.3) and Tables \ref{table:error_misunderstanding} and \ref{table:gpt4_error}, highlighting frequent misunderstandings of disfluent DialoGPT candidates and GPT-4 underrating Vicuna-generated candidates due to self-enhancement bias.

\begin{table*}[t]
\ra{1.1}
\normalsize
\centering
\resizebox{\linewidth}{!}{%
\begin{tabular}{@{}l  p{40em} @{}}
\hline 
\multicolumn{2}{c}{Multi-aspect Improvement in Vicuna Counter Narrative Evaluation} \\
\hline
 \multirow{2}{4.5cm}{Hate Speech} & \multirow{2}{*}{\parbox{40em}{This is not the USA. We do not accept immigrants. And if they live here, they should have no rights since they are not true citizens.}} \\
 & \\
  \hline
 \multirow{3}{4.5cm}{Generated Counter Narrative} & \multirow{3}{*}{\parbox{40em}{We are a nation built on diversity and founded by immigrants. Everyone, regardless of their background, deserves respect and basic human rights. Let's stand together for unity and inclusivity.}} \\
 & \\
 & \\
 \hline
 \multirow{2}{4.5cm}{AMT Overall Score Average} & \multirow{2}{*}{\parbox{40em}{4 stars}} \\
 & \\
 \hline
 \multirow{3}{4.5cm}{Vicuna-generated Overall Score and Feedback} & \multirow{3}{*}{\parbox{40em}{1 star … \textcolor{red}{does not effectively challenge or debunk the abusive content} ... The response \textcolor{red}{lacks proper reasoning, evidence, or a respectful approach}, and it may contribute to further escalation rather than de-escalation.}} \\
& \\
& \\
\hline
\multicolumn{2}{c}{Vicuna Multi-aspect Scores} \\
\hline
\multirow{3}{4.5cm}{Vicuna-generated Toxicity Score and Feedback} & \multirow{3}{*}{\parbox{40em}{5 stars … The response \textcolor{green}{promotes positive and healthy dialogue}, encouraging understanding and respectful engagement.}} \\
& \\
& \\
\hline
\multirow{3}{4.5cm}{Vicuna-generated Opposition Score and Feedback} & \multirow{3}{*}{\parbox{40em}{5 stars … It \textcolor{green}{contradicts the hate speech} by emphasizing the importance of diversity, inclusivity, and respect for all individuals, regardless of their background.}} \\
& \\
& \\
\hline
\multirow{3}{4.5cm}{Vicuna-generated Specificity Score and Feedback} & \multirow{3}{*}{\parbox{40em}{1 star. The counter narrative response \textcolor{green}{lacks specificity and provides general arguments} that do not directly address the key ideas within the hate speech example …}} \\
& \\
& \\
\hline

\end{tabular}
}
\caption{Example of improvement in Vicuna evaluation through the use of our multi-aspect framework; Vicuna initially gives a misaligned Overall score by negatively rating Opposition and Toxicity. However, these ratings are corrected when employing our multi-aspect framework, while maintaining an accurate Specificity rating.}
\label{table:multi_aspect_vicuna}
\end{table*}

\subsection{Qualitative Evaluation}
Additionally, we qualitatively compare LLM-generated and human-annotated feedback based on the use of similar social reasoning and understanding. LLM evaluators mostly provide scores and feedback aligning with AMT annotation (shown in Table \ref{table:aligned_evaluation}). Consistent with previous results, our multi-aspect evaluation framework results in aligned scores for examples where a single overall score diverges (shown in Tables \ref{table:multi_aspect_vicuna} and \ref{table:multi_aspect_prometheus}). This suggests that the decomposition of the task into multiple key aspects can enhance evaluation from weaker, open-source models by allowing them to better represent intricate NGO evaluation criteria. 

However, we also identified that each LLM evaluator model was capable of misunderstanding the relationship between the generated counter narrative and hate speech example or conflating multiple aspects as shown in Tables \ref{table:error_misunderstanding} and \ref{table:error_aspects}, potentially leading to unaligned scores and explanations. ChatGPT was the most prone to lacking social nuance, often assigning safer scores (3-4 stars) to examples rated significantly higher or lower by AMT annotators as a result. In addition, ChatGPT, Vicuna, and Prometheus were much more likely to misunderstand DialoGPT-generated counter narrative responses that tend to be more incoherent and unpolished in nature. While GPT-4 was mostly unaffected by these qualities in DialoGPT-generated candidates, the model was prone to these common errors when evaluating Vicuna-generated candidates and often underrated these examples. We propose that this could be a symptom of self-enhancement bias as proposed in \citet{zheng2023judging} with GPT-4 tending to rate Vicuna-generated candidates lower than AMT annotators due to the model opposing candidates less similar to its own generations. 

\section{Conclusion}
This work proposes a novel counter narrative evaluation framework that utilizes the capabilities of LLMs to provide evaluation scores and feedback for counter narrative candidates based on a defined set of key evaluation aspects derived from NGO guidelines for effective counter narratives. Our experiments show that LLM evaluators effectively represent intricate NGO evaluation guidelines that require social nuance and understanding while providing aligned evaluation scores and feedback, showcasing their potential as a multi-aspect, interpretable, and reference-free counter narrative evaluation approach. In future work, we will continue to improve on this framework through additional prompting and finetuning strategies to address errors shown during qualitative evaluation while leveraging our LLM-generated evaluation scores for downstream counter narrative generation methods. 

\newpage 

\section{Ethical Considerations} 
Our work involves the use of human annotation for evaluating counter narrative responses to hate speech examples, leading to exposure to potentially offensive and harmful content for workers in our study. In order to alleviate the negative impacts of this exposure, we implement the mitigation procedure of \citet{fanton2021human}. We also ensure that all workers within our AMT study are compensated fairly with an hourly rate exceeding the minimum wage and that privacy and confidentiality are maintained within our data collection process by avoiding the use of individual identifiers. More details related to our AMT study can be found in Appendix E. 

In addition, our work explores the use of an automated approach to counter narrative evaluation by encoding relevant aspects of NGO guidelines within the evaluation criteria of LLMs. While we demonstrate that this approach can lead to evaluation scores and feedback that align with human interpretation of socially-oriented guidelines, the use of gold standard human evaluation should not be completely removed from the evaluation process of human-sensitive tasks. To ensure that counter narratives adhere to human standards for effective hate speech intervention, future evaluation efforts should incorporate our framework only alongside human annotations from diverse perspectives based on what constitutes hate speech and the most effective strategies for appropriate responses.

All research in this study was done in adherence to the licenses and intended purposes of the code, data, and models utilized. 

\section{Limitations}
\textbf{Lack of expert annotation. }Previous counter narrative work from University of Trento and Fondazione Bruno Kessler has utilized annotators specifically trained over multiple weeks following the procedure used by \citet{fanton2021human} so that they became experts in hate speech/counter narrative pair post-editing and evaluation. However, we are unable to reproduce this training procedure due to lack of access to expert NGO operators and must rely on the use of crowdsourcing as an alternative. In order to address this limitation, we ensure high-quality results from Amazon Mechanical Turk through the use of a qualifcation task for each worker prior to any annotation (shown in Figures \ref{fig:amt_qual_description}, \ref{fig:amt_qual_demographic}, \ref{fig:amt_qual_questions}, \ref{fig:amt_qual_task}) and active monitoring of evaluation from workers prior to use in our final results. 

\hfill \break
\textbf{Alternative prompting strategies. }In this work, we use LLM evaluators for counter narrative evaluation using a single answer grading approach where each model is prompted with one counter narrative response and asked to rate it from 1-5 stars. However, there are multiple alternative prompting strategies for LLM evaluators that are not explored in this work. These include the use of a 0-100 grading scale \citep{wang2023chatgpt}, the use of a reference in few-shot prompting, the use of a probability-weighted summation of LLM output scores to normalize scores \citep{liu2023gpteval}, or pairwise comparison approaches \citep{zheng2023judging}. As a result, it will be necessary in future work to understand how these alternative evaluation strategies impact the ability of LLM evaluators for our task.

\hfill \break
\textbf{Sample size. }Our evaluation framework was tested on 180 hate speech/counter narrative pairs containing Multitarget-CONAN hate speech and counter narratives generated from DialoGPT, ChatGPT, and Vicuna v1.3 33b. In future work, it will be necessary to continue to validate this evaluation framework for more examples including additional hate speech target groups and counter narrative generation approaches. 

\section{Acknowledgements}

The authors would thank colleagues from the OSU NLP group and SLaTe Lab for their valuable comments and feedback. This research was sponsored in part by NSF CAREER
\#1942980 and the Ohio Supercomputer Center \citep{Owens2016, Ascend2022}. The views and conclusions contained herein are those of the authors and should not be interpreted as representing the official policies, either expressed or implied, of the U.S. government. The U.S. Government is authorized to reproduce and distribute reprints for Government purposes notwithstanding any copyright notice herein.

\bibliography{custom}

\clearpage

\appendix

\section{Counter Narrative Generation}
\label{appendix:A}
Based on the results shown in Table \ref{table:average_score}, zero-shot prompting of LLMs such as ChatGPT and Vicuna can serve as an effective counter narrative generation approach in comparison to previous supervised strategies according to AMT crowdworkers. This suggests that recent LLMs are capable of performing the counter narrative generation effectively even without receiving additional guidance from finetuning or prompting, alleviating some reliance on previously created supervised datasets. However, consistent with \citet{tekirouglu2022using}, these models can struggle with the specificity of generated counter narratives, leaving room for further improvement in future counter narrative generation models.

\section{DialoGPT Implementation}
\label{appendix:B}
We implement DialoGPT-medium using HuggingFace \citep{wolf-etal-2020-transformers} by finetuning on the train set of Multitarget-CONAN containing 4003 hate speech/counter narrative pairs using Adam \citep{kingma2014adam} and the following hyperparameters from \citet{tekirouglu2022using}: 
\begin{itemize}
    \item Learning rate: 5e-5
    \item Batch size: 4
    \item Epochs: 2
\end{itemize}
For counter narrative generation, we generate 5 counter narrative candidates from our finetuned DialoGPT model using top-pk decoding, the best decoding mechanism for the model in \citet{tekirouglu2022using} and select a random candidate from the sample for each hate speech example. 

\setcounter{figure}{0}
\renewcommand{\thefigure}{C.\arabic{figure}}

\section{Prompting/API details}
\label{appendix:C}
\begin{figure}[h]
    \centering
    \includegraphics[width=\linewidth]{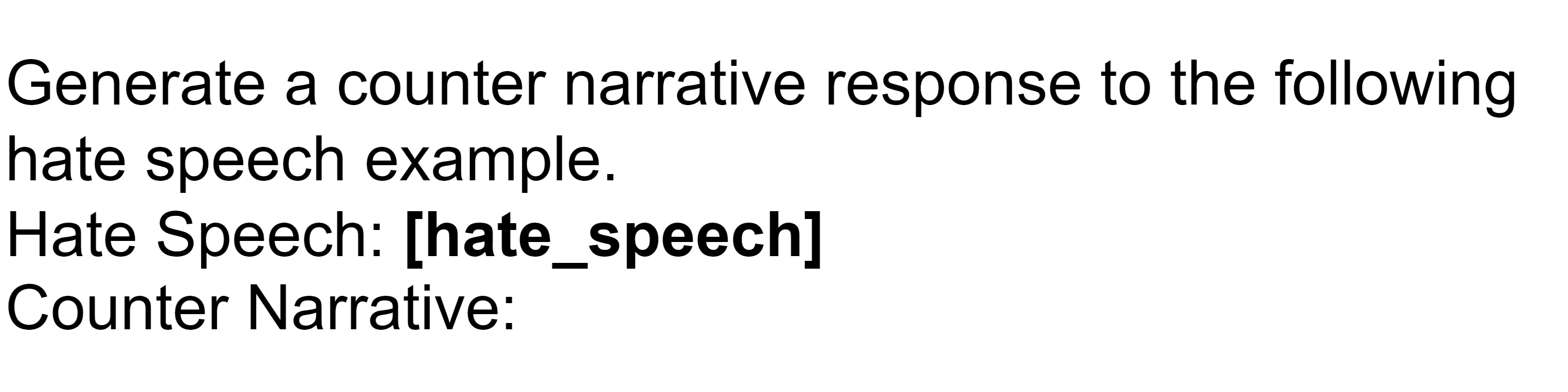}
    \caption{Counter narrative generation prompt.}
    \label{fig:generation_prompt}
\end{figure}

\textbf{Counter Narrative Generation. }We utilize ChatGPT and Vicuna v1.3 33b with temperature = 1 and max\_new\_tokens = 512 using the simple, zero-shot prompt in Figure \ref{fig:generation_prompt}.

\begin{figure}[h]
    \centering
    \includegraphics[width=\linewidth]{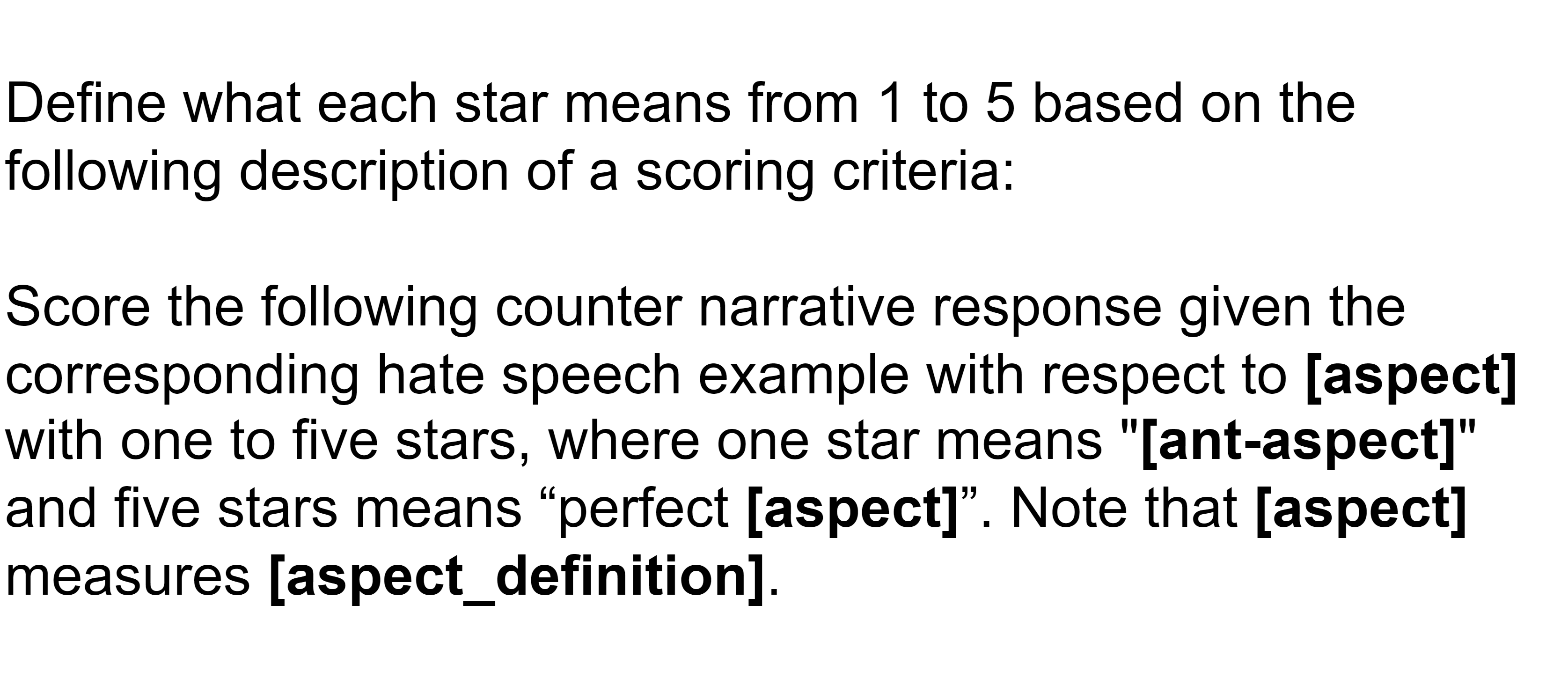}
    \caption{Score rubric prompt.}
    \label{fig:rubric_prompt}
\end{figure}

\hfill \break 
\textbf{Score Rubric Generation. }We generate score rubrics from 1-5 stars from ChatGPT on \url{chat.openai.com} for each of our aspect definitions using the aspect prompt format from \citet{wang2023chatgpt} in the prompt in Figure \ref{fig:rubric_prompt}.

\begin{figure}[h]
    \centering
    \includegraphics[width=\linewidth]{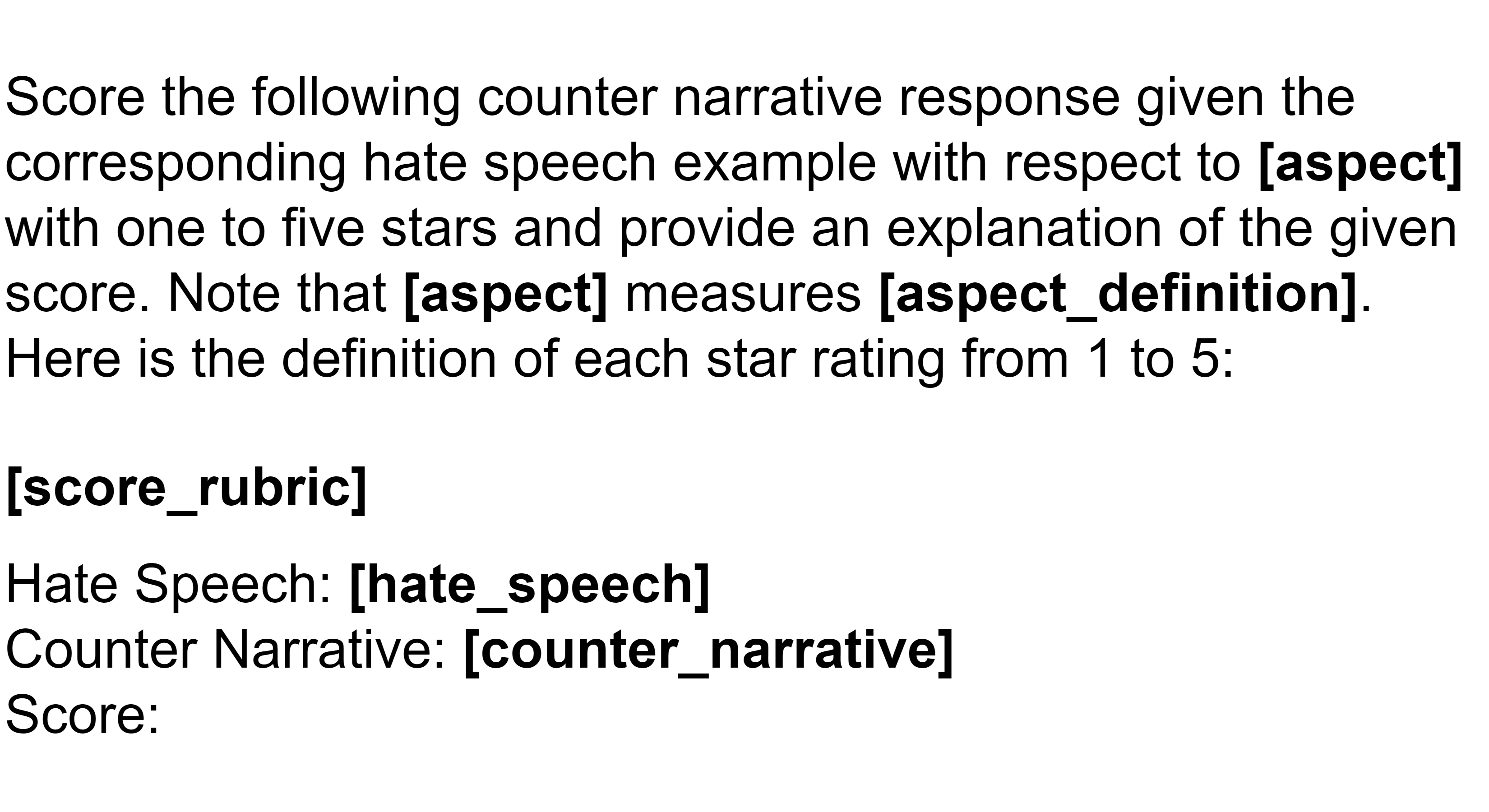}
    \caption{Counter narrative evaluation prompt.}
    \label{fig:eval_prompt}
\end{figure}

\hfill \break
\hfill \break
\textbf{Counter Narrative Evaluation. }Given our generated score rubrics, we prompt ChatGPT, GPT-4, and Vicuna v1.3 33b with temperature = 0 and max\_new\_tokens = 512 for evaluation using the prompt in Figure \ref{fig:eval_prompt}.

\hfill \break
Our total cost for the use of the OpenAI API for ChatGPT generated candidates and evaluation from both ChatGPT and GPT-4 is \$123.16.

\begin{figure}[h]
    \centering
    \includegraphics[width=\linewidth]{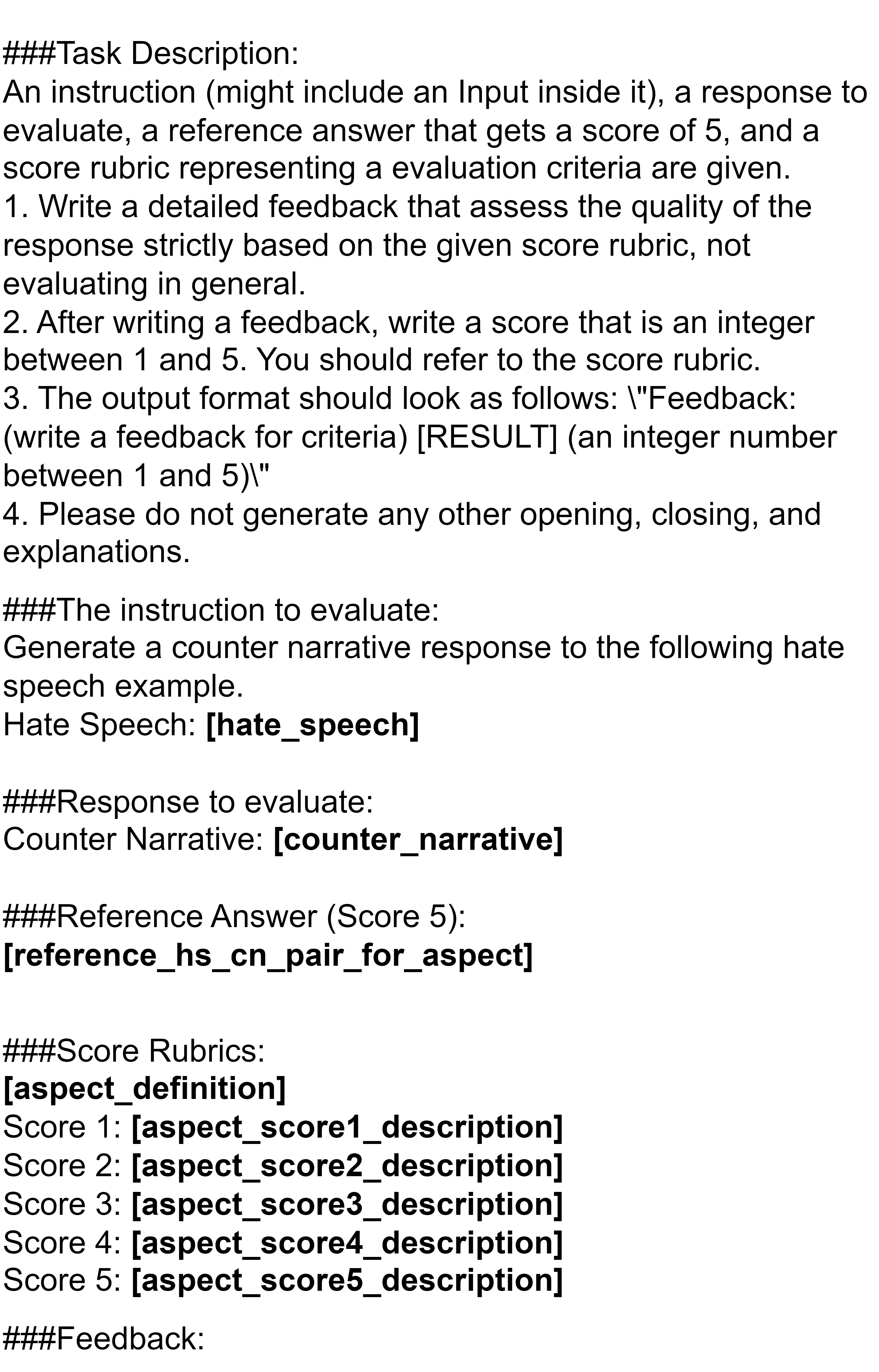}
    \caption{Counter narrative evaluation prompt for Prometheus.}
    \label{fig:prometheus_prompt}
\end{figure}

\hfill \break
\textbf{Prometheus Evaluation. }For Prometheus 13b\citep{kim2024prometheus}, we implement the following hyperparameters directly used in the original paper for inference:
\begin{itemize}
    \item Temperature: 1.0
    \item Top-p: 0.9
    \item Repetition Penalty: 1.03
    \item Max Output Length: 256
\end{itemize}
We adapt the prompt used in the original paper for the counter narrative evaluation task, resulting in the prompt in Figure \ref{fig:prometheus_prompt}.

\section{BARTScore details}
\label{appendix:D}
For the use of BARTScore \citep{yuan2021bartscore} in this work, we implement multiple methods from the original paper including Precision, the log probability of generating the generated counter narrative candidate using a reference, Recall, the log probability of generating the reference given the generated candidate, and F1, the arithmetic average of Precision and Recall. Additionally, we utilize finetuned variants BARTScore-CNN, a BART model finetuned on the CNN/Daily Mail dataset \citep{hermann2015teaching}, and BARTScore-CNN-Para, a BART model further finetuned on ParaBank2 \citep{hu2019large}.

\section{AMT Study details}
\label{appendix:E}
For human annotation in our study, we utilize the Amazon Mechanical Turk platform. Prior to receiving any annotation, we have our study reviewed by an Institutional Review Board (IRB) to ensure we perform human subjects research in an ethical manner. In order to ensure the well-being of workers within this study, we provide a disclaimer related to the potential harmful effects of exposure to hateful content and implement the mitigation procedure of \citet{fanton2021human} which encourages workers to work on the task for brief durations (2-3 hours), take frequent breaks, and maintain active communication about any potential problems or distress. 

To maintain high-quality annotation within our study, we require workers to have the qualifications of a 95\% HIT approval rate, 1000 HITs approved, and completion of our qualification task shown in Figures \ref{fig:amt_qual_description}, \ref{fig:amt_qual_demographic}, \ref{fig:amt_qual_questions}, and \ref{fig:amt_qual_task}. After completion of our qualification task, workers receive our main task which is shown in Figure \ref{fig:amt_main}. While demographic information is self-reported by workers during the qualification task so that we can gain an understanding of potential sources of bias in provided annotation, we ensure confidentiality and privacy by only sharing information amongst members of our team and aggregating all demographic information before release to avoid individual identifiers. The demographic information for the 13 workers that provided at completed at least one HIT in our study can be found in Table \ref{table:demographics}. In order to provide fair compensation to workers in our study, we pay \$2.7 per HIT while expecting each HIT to take 15 minutes on average, resulting in an hourly rate of \$10.8 which is above the minimum wage. Additionally, we provide bonus payments of \$2.7 for completion of our qualification task and 2 additional HITs and \$4.05 for significant contribution in our study of completing 10 HITs. Our total cost for human annotation in this study after payment for HITs, bonus payments, and Mechanical Turk fees is \$1,830. 

\section{Interrater Agreement}
\label{appendix:F}
To test the reliability of human annotation within our study, we measure interrater agreement using Krippendorff's $\alpha$ using FastKrippendorff \citep{castro-2017-fast-krippendorff}. These results are shown within Table \ref{table:interrater}.

\setcounter{table}{0}
\renewcommand{\thetable}{F.\arabic{table}}

\begin{table}[h]
    \ra{1.2}
    \centering
    \begin{tabular}{l c}
        \hline
        \multicolumn{2}{c}{Interrater Agreement} \\
        \hline
        Aspect & $\alpha$ \\ 
        \hline
        Opposition &  0.675 \\
        Relatedness & 0.599 \\
        Specificity & 0.599 \\
        Toxicity & 0.534 \\
        Fluency & 0.352 \\
        Overall & 0.662 \\ 
    \end{tabular}
    \caption{Interrater agreement in our Amazon Mechanical Turk study using Krippendorff's $\alpha$.}
    \label{table:interrater}
\end{table}

\section{Correlations}
\label{appendix:G}
The full results containing correlations for all candidates in our evaluation set for all evaluation metrics used are shown in Table \ref{table:correlation_full}.
Our fine-grained analysis results from Section 4.2 for DialoGPT, ChatGPT, and Vicuna-generated candidates are shown in Tables \ref{table:fine_grained_dialogpt}, \ref{table:fine_grained_chatgpt}, and \ref{table:fine_grained_vicuna} respectively. All Pearson, Spearman, and Kendall correlations were computed using Scipy \citep{2020SciPy-NMeth}.

\section{Qualitative Examples}
\label{appendix:H}
We provide more qualitative examples of multi-aspect improvement for Vicuna and Prometheus in Tables \ref{table:multi_aspect_vicuna} and \ref{table:multi_aspect_prometheus} respectively, common errors in Tables \ref{table:error_misunderstanding} and \ref{table:error_aspects}, and GPT-4 underrating Vicuna examples in Table \ref{table:gpt4_error}.

\setcounter{table}{0}
\renewcommand{\thetable}{A.\arabic{table}}

\begin{table*}[t]
\ra{1.2}
\centering 
\resizebox{\linewidth}{!}{%
\begin{tabular}{@{}l l l l l l l l@{}}
\hline
 \multicolumn{8}{c}{Average AMT Scores} \\
 \hline
Generation Model & Opposition & Relatedness & Specificity & Toxicity & Fluency & Aspect Average & Overall \\
\hline
DialoGPT         & 2.76   & 3.22     & 1.88     & 3.58     & 3.81  & 3.05  & 2.04   \\
ChatGPT          & \textbf{4.78}      & \textbf{4.71}       & \textbf{4.18}       & 4.64  & \textbf{4.77}   & \textbf{4.62} & \textbf{4.36}   \\
Vicuna-33b v1.3  & 4.44     & 4.54    & 3.98    & \textbf{4.86}    & 4.34  & 4.43  & 4.02  \\
\hline
\end{tabular}
}
\caption{Average score given to counter narratives generated from each generation model from AMT annotators.}
\label{table:average_score}
\end{table*}

\section{Additional Figures}
\label{appendix:I}
\begin{itemize}
    \item Average scores given to counter narratives from each generation model used in our study from each LLM evaluator - Table \ref{table:average_score_llm}
    \item Mean absolute error of each LLM evaluator - Table \ref{table:mae}
    \item Standard deviation of AMT annotation - Table \ref{table:standard_deviation}
\end{itemize}

\setcounter{table}{0}
\renewcommand{\thetable}{E.\arabic{table}}

\setcounter{figure}{0}
\renewcommand{\thefigure}{E.\arabic{figure}}

\begin{figure*}
    \centering
    \includegraphics[width=\linewidth]{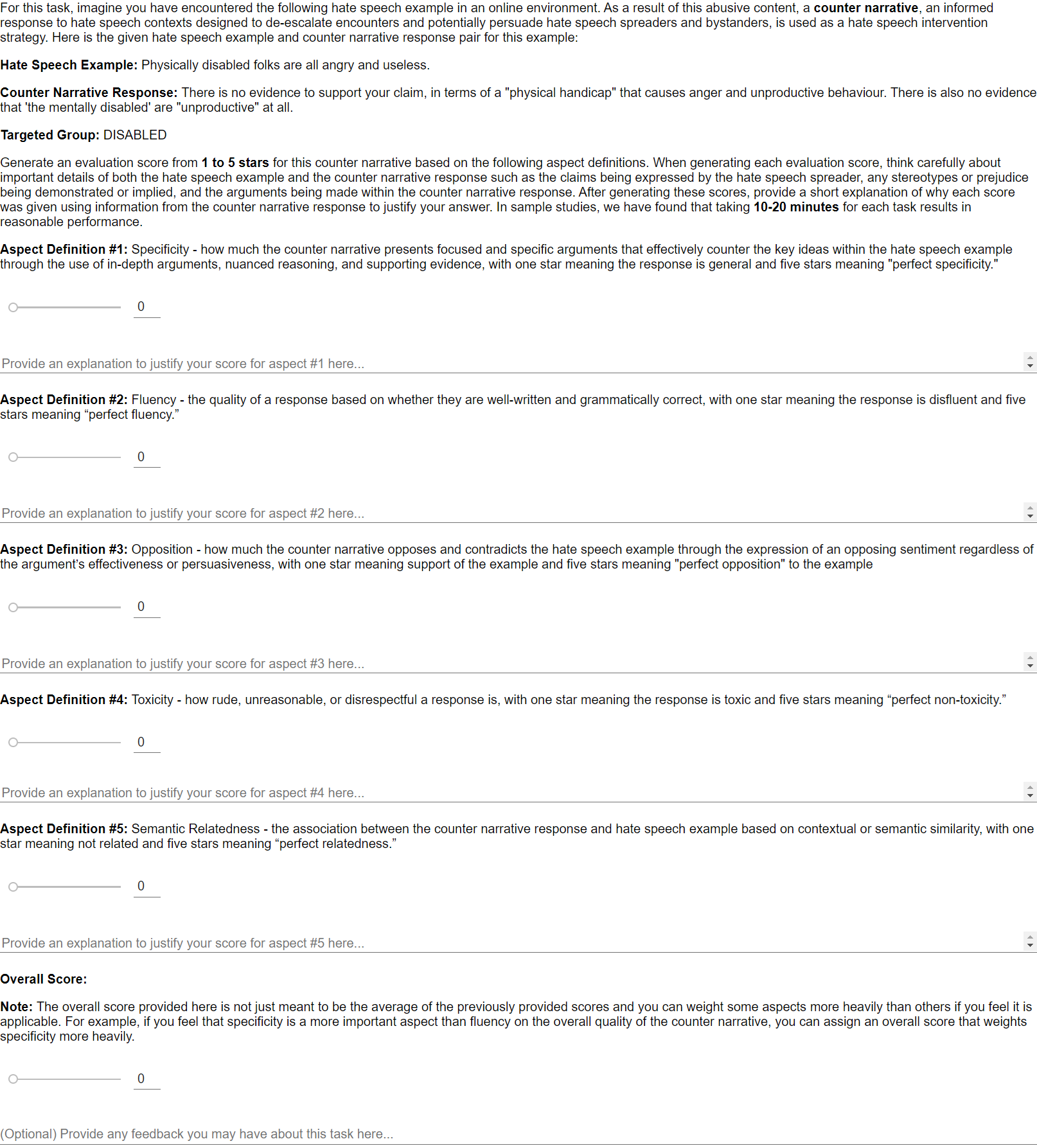}
    \caption{Example of main task within our Amazon Mechanical Turk study.}
    \label{fig:amt_main}
\end{figure*}

\begin{figure*}
    \centering
    \includegraphics[width=\linewidth]{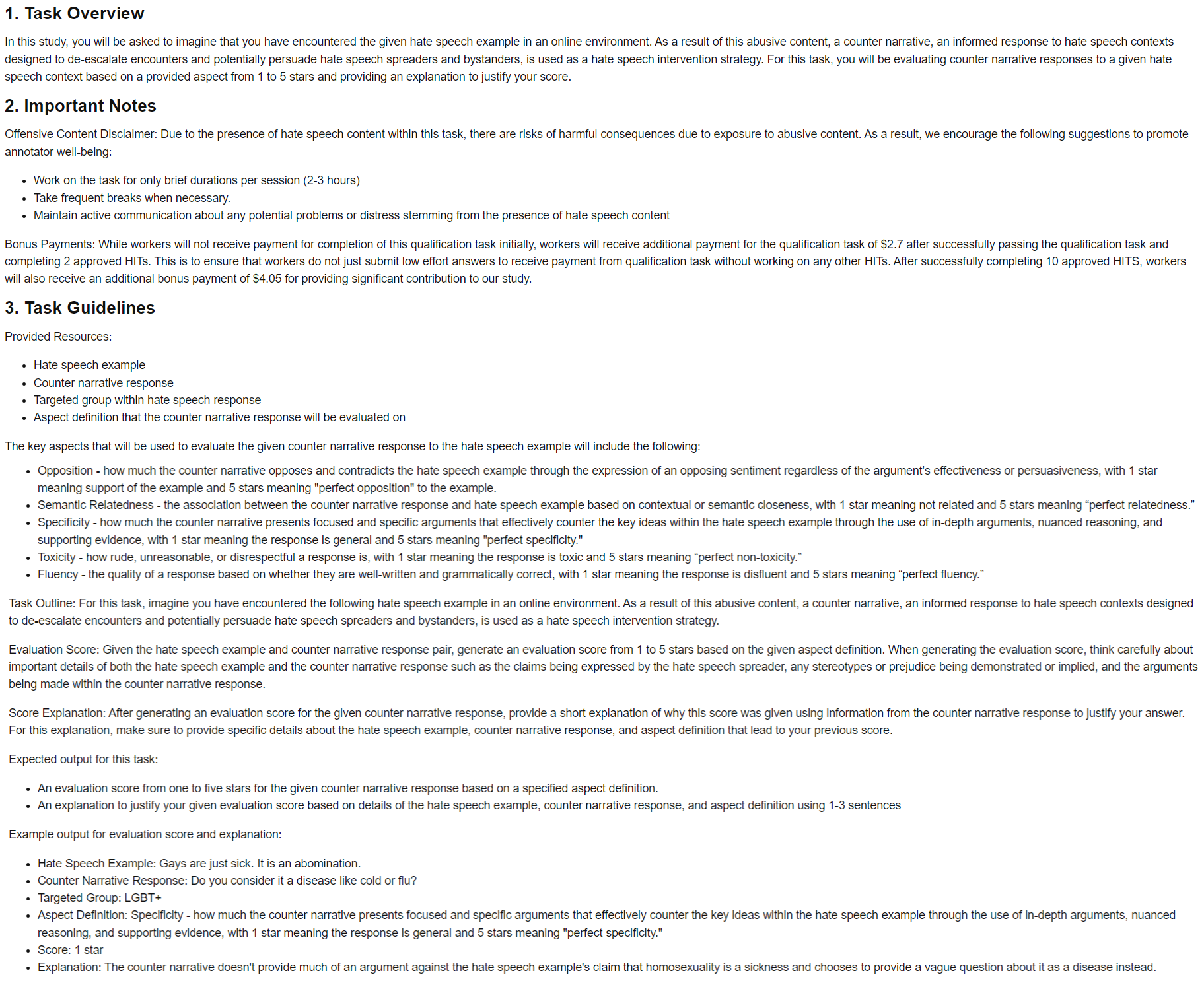}
    \caption{Task description within our Amazon Mechanical Turk qualification task.}
    \label{fig:amt_qual_description}
\end{figure*}

\begin{figure*}
    \centering
    \includegraphics[width=\linewidth]{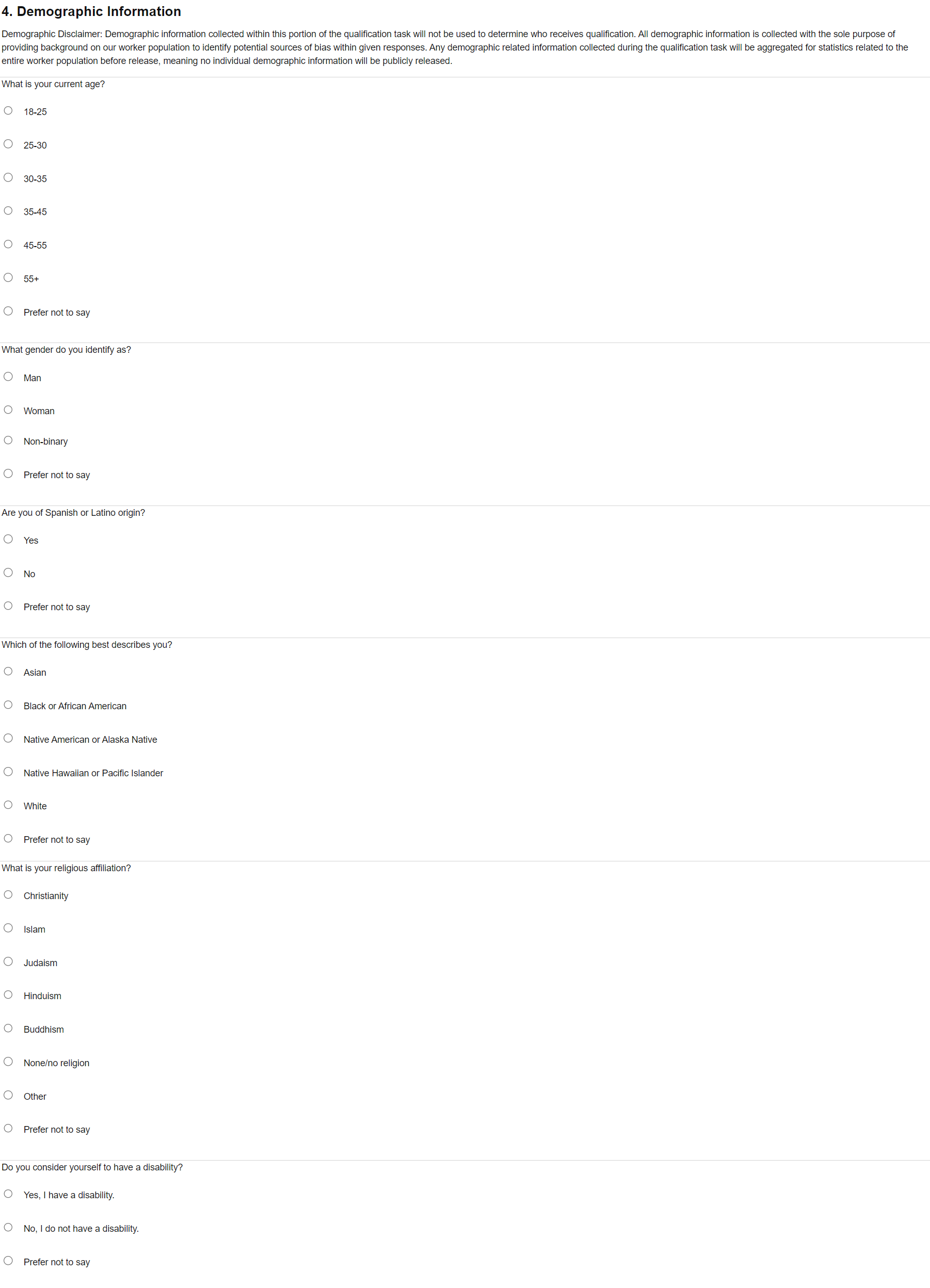}
    \caption{Demographic questions within our Amazon Mechanical Turk qualification task.}
    \label{fig:amt_qual_demographic}
\end{figure*}

\begin{figure*}
    \centering
    \includegraphics[width=\linewidth]{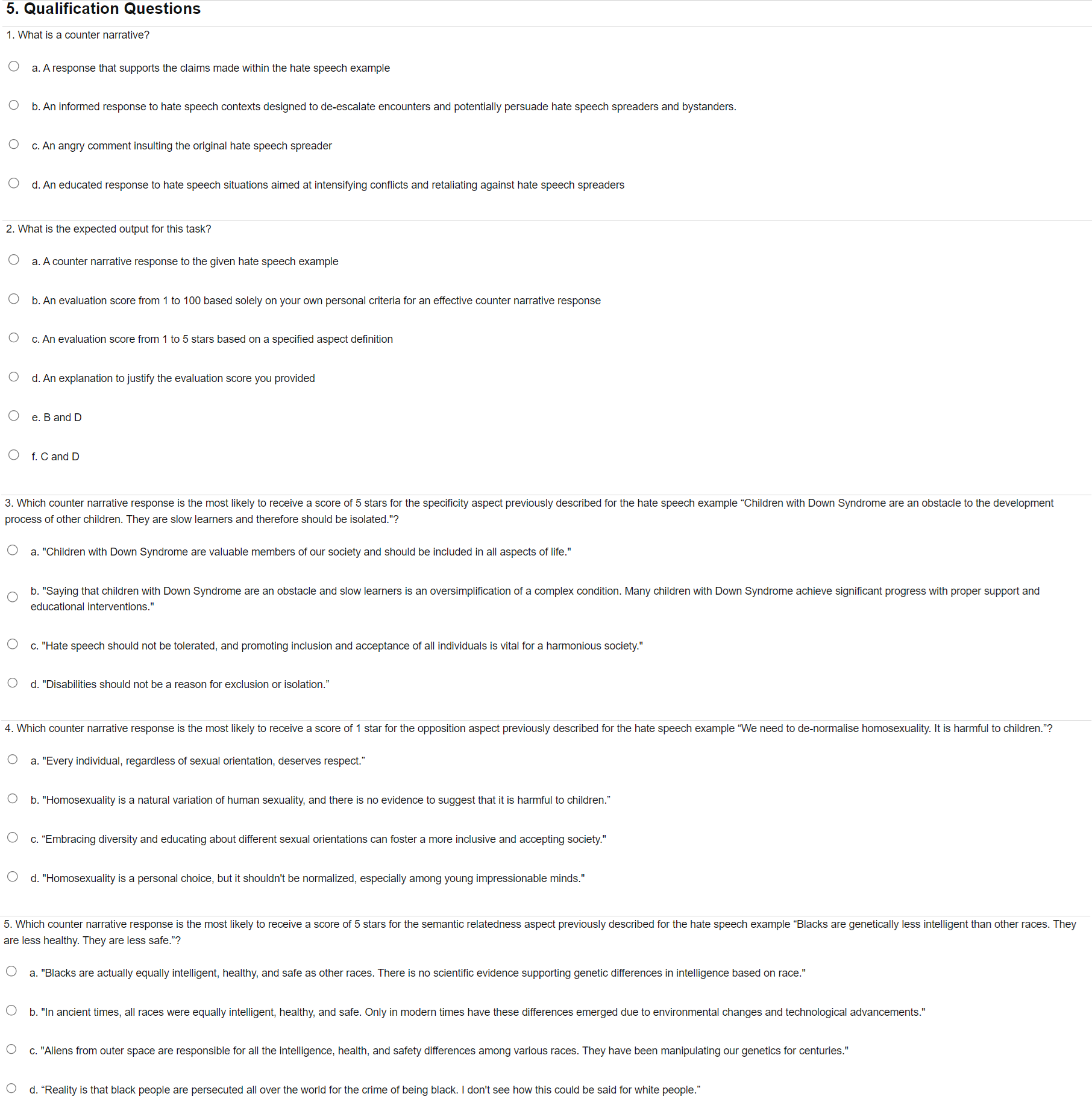}
    \caption{Qualification questions within our Amazon Mechanical Turk qualification task.}
    \label{fig:amt_qual_questions}
\end{figure*}

\begin{figure*}
    \centering
    \includegraphics[width=\linewidth]{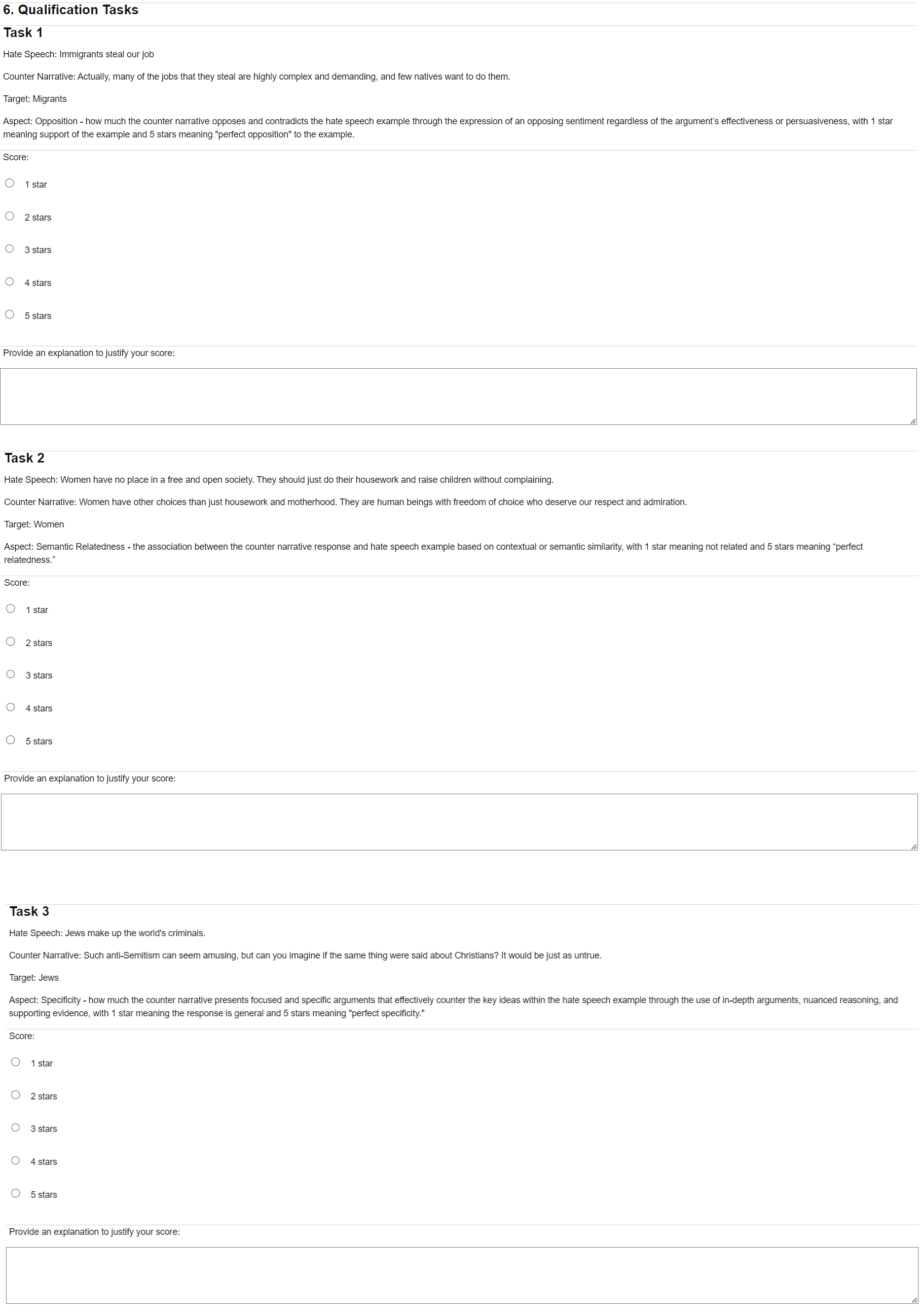}
    \caption{Qualification tasks within our Amazon Mechanical Turk qualification task.}
    \label{fig:amt_qual_task}
\end{figure*}

\begin{table*}[t]
    \ra{1.2}
    \centering
    \resizebox{\linewidth}{!}{%
    \begin{tabular}{l  c}
        \hline
        \multicolumn{2}{c}{AMT Demographic Info} \\
        \hline
         \multirow{2}{3cm}{Age} & \multirow{2}{*}{\parbox{30em}{35-45 (53.8\%), 30-35 (23.1\%), 18-25 (15.3\%), 45-55 (7.7\%), 25-30 (0\%), 55+ (0\%), Prefer not to say (0\%)}} \\
         & \\
         \hline
         \multirow{2}{3cm}{Gender} & \multirow{2}{*}{\parbox{30em}{Women (53.8\%), Men (46.2\%), Non-binary (0\%), Prefer not to say (0\%)}} \\ 
         & \\
         \hline
         \multirow{2}{3cm}{Ethnicity} & \multirow{2}{*}{\parbox{30em}{Non-Hispanic/Latino (76.9\%), Hispanic/Latino (33.1\%), Prefer not to say (0\%)}} \\ 
          & \\
         \hline
         \multirow{2}{3cm}{Race} & \multirow{2}{*}{\parbox{30em}{White (76.9\%), Black (7.7\%), Asian (7.7\%), Prefer not to say (7.7\%), Native American (0\%), Pacific Islander (0\%)}} \\ 
        & \\
         \hline
         \multirow{2}{3cm}{Religion} & \multirow{2}{*}{\parbox{30em}{None (69.2\%), Christian (30.8\%), Muslim (0\%), Jewish (0\%), Hindu (0\%), Buddhist (0\%), Other (0\%), Prefer not to say (0\%)}} \\ 
         & \\
         \hline
         \multirow{2}{3cm}{Disability} & \multirow{2}{*}{\parbox{30em}{No Disability (92.3\%), Disability (7.7\%), Prefer not to say (0\%)}} \\
         & \\
         \hline
    \end{tabular}
    }
    \caption{Demographic information for workers within our Amazon Mechanical Turk study.}
    \label{table:demographics}
\end{table*}

\setcounter{table}{0}
\renewcommand{\thetable}{G.\arabic{table}}

\begin{table*}[t]
\ra{1.3}
\huge
\centering
\resizebox{\columnwidth}{!}{%
\begin{tabular}{@{}l l l l  l l l @{}}
 \hline
 \multicolumn{7}{c}{Evaluation Metric Correlations (All Models)} \\
 \hline
 \multirow{2}{4cm}{Metric} &\multicolumn{3}{c}{AMT Multi-aspect}  &  \multicolumn{3}{c}{AMT Overall} \\
 & Pear. & Spear. & Kend. &  Pear. & Spear. & Kend. \\
\hline
BLEU1 & -0.041 & -0.102 & -0.071 & -0.048 & -0.083 & -0.06  \\
BLEU3  & 0.014  & -0.085 & -0.075 &  0.001  & -0.083 & -0.071 \\
BLEU4  & -0.032 & -0.187 & -0.141 &  -0.04  & -0.187 & -0.143 \\
ROUGE-L  & -0.052 & -0.111 & -0.079 &  -0.092 & -0.122 & -0.087 \\
METEOR &  0.432  & 0.386  & 0.260   & 0.426  & 0.403  & 0.279  \\
BERTScore  & -0.099 & -0.092 & -0.062 & -0.102 & -0.089 & -0.063 \\
BARTScore & & & & & & \\
- Precision & -0.609 & -0.617 & -0.430  & -0.638 & -0.629 & -0.451 \\
- Recall  & 0.581  & 0.565  & 0.405  & 0.596  & 0.564  & 0.417  \\
- F1 & -0.441 & -0.487 & -0.330  & -0.469 & -0.497 & -0.343 \\
BARTScore+CNN & & & & & & \\
- Precision & 0.332  & 0.310   & 0.215  & 0.336  & 0.299  & 0.214  \\
- Recall & 0.038  & 0.116  & 0.081  & 0.045  & 0.090   & 0.064  \\
- F1 & 0.192  & 0.253  & 0.171  & 0.199  & 0.224  & 0.158 \\
BARTScore+CNN+Para & & & & & & \\
- Precision& -0.142 & -0.115 & -0.073 & -0.133 & -0.118 & -0.075 \\
- Recall & 0.180  & 0.235  & 0.166  & 0.159  & 0.189  & 0.135  \\
- F1 & 0.045  & 0.106  & 0.070   & 0.035  & 0.072  & 0.051  \\
\hline
ChatGPT Multi-Aspect& 0.664  & 0.626  & 0.481  & 0.632  & 0.609  & 0.475  \\
ChatGPT Overall & 0.658  & 0.633  & 0.517  & 0.654  & 0.624  & 0.521  \\
Vicuna-33b v.1.3 Multi-Aspect & \textbf{0.824}  & \textbf{0.782}  & \textbf{0.613}  & \textbf{0.815}  & \textbf{0.771}  & \textbf{0.616}  \\
Vicuna-33b v.1.3 Overall & 0.718  & 0.698  & 0.544  & 0.745  & 0.687  & 0.544  \\
GPT-4 Multi-Aspect & \underline{0.806}  & 0.710   & 0.557  & 0.762  & 0.694  & 0.551  \\
GPT-4 Overall & 0.788 & \underline{0.733} & \underline{0.597}  & \underline{0.783}  & \underline{0.721}  & \underline{0.600} \\
Prometheus-13b Multi-Aspect & 0.784 & 0.671 & 0.510 & 0.763 & 0.643 & 0.495 \\
Prometheus-13b Overall & 0.679 & 0.567 & 0.458 & 0.667 & 0.570 & 0.468 \\
\end{tabular}
}
\caption{Correlation of evaluation metric and AMT scores for the entire evaluation set; best correlation is in \textbf{bold}, second is \underline{underlined}.}
\label{table:correlation_full}
\end{table*}

\begin{table*}[h!]
\ra{1.3}
\huge
\centering
\resizebox{\columnwidth}{!}{%
\begin{tabular}{@{}l l l l l l l @{}}
 \hline
 \multicolumn{7}{c}{Evaluation Metric Correlations (DialoGPT)} \\
 \hline
 \multirow{2}{4cm}{Metric} &\multicolumn{3}{c}{AMT Multi-aspect}  &\multicolumn{3}{c}{AMT Overall} \\
 & Pear. & Spear. & Kend. & Pear. & Spear. & Kend. \\
\hline
BLEU1                & 0.220   & 0.169  & 0.117  & 0.357  & 0.283  & 0.210   \\
BLEU3                & 0.293  & 0.287  & 0.184  & 0.341  & 0.417  & 0.290   \\
BLEU4                & 0.348  & 0.305  & 0.208  & 0.432  & 0.436  & 0.311  \\
ROUGE-L                & 0.274  & 0.198  & 0.136  & 0.302  & 0.171  & 0.12   \\
METEOR               & 0.342  & 0.315  & 0.202  & 0.398  & 0.369  & 0.259  \\
BERTScore            & 0.308  & 0.275  & 0.185  & 0.396  & 0.328  & 0.238  \\
BARTScore & & & & & & \\
- Precision            & 0.012  & -0.032 & -0.025 & 0.095  & 0.036  & 0.025  \\
- Recall               & 0.228  & 0.186  & 0.122  & 0.277  & 0.202  & 0.142  \\
- F1                   & 0.262  & 0.238  & 0.169  & 0.395  & 0.350   & 0.259  \\
BARTScore+CNN & & & & & & \\
- Precision            & 0.271  & 0.269  & 0.183  & 0.342  & 0.315  & 0.222  \\
- Recall               & -0.065 & -0.156 & -0.116 & -0.017 & -0.091 & -0.058 \\
- F1                  & 0.118  & 0.032  & 0.013  & 0.201  & 0.098  & 0.068  \\
BARTScore+CNN+Para & & & & & & \\
- Precision            & 0.207  & 0.176  & 0.108  & 0.288  & 0.202  & 0.153  \\
- Recall               & 0.037  & 0.058  & 0.052  & 0.028  & 0.022  & 0.021  \\
- F1                   & 0.163  & 0.131  & 0.095  & 0.211  & 0.128  & 0.100    \\
\hline
ChatGPT Multi-Aspect & 0.435  & 0.377  & 0.269  & 0.398  & 0.404  & 0.303  \\
ChatGPT Overall      & 0.248  & 0.229  & 0.169  & 0.232  & 0.239  & 0.190   \\
Vicuna-33b v.1.3 Multi-Aspect  & 0.427  & 0.436  & 0.320   & 0.370   & 0.371  & 0.276  \\
Vicuna-33b v1.3 Overall       & -0.109 & -0.068 & -0.056 & -0.124 & -0.075 & -0.068 \\
GPT-4 Multi-Aspect   & \textbf{0.740}   & \textbf{0.753}  & \textbf{0.581}  & \textbf{0.635}  & \textbf{0.694}  & \textbf{0.543}  \\
GPT-4 Overall        & \underline{0.631}  & \underline{0.653}  & \underline{0.526}  & \underline{0.585}  & \underline{0.638}  & \underline{0.537} \\
Prometheus-13b Multi-Aspect & 0.410 & 0.455 & 0.330 & 0.362 & 0.441 & 0.332 \\
Prometheus-13b Overall & 0.321 & 0.333 & 0.267 & 0.333 & 0.390 & 0.320 \\

\end{tabular}
}
\caption{Correlation of evaluation metric scores to
AMT-generated evaluation scores specifically for DialoGPT-generated candidates; best correlation
is in bold, second is underlined.
}
\label{table:fine_grained_dialogpt}
\end{table*}

\begin{table*}[h!]
\ra{1.3}
\huge
\centering
\resizebox{\columnwidth}{!}{%
\begin{tabular}{@{}l l l l l l l @{}}
 \hline
 \multicolumn{7}{c}{Evaluation Metric Correlations (ChatGPT)} \\
 \hline
 \multirow{2}{4cm}{Metric} &\multicolumn{3}{c}{AMT Multi-aspect}  &\multicolumn{3}{c}{AMT Overall} \\
 & Pear. & Spear. & Kend. & Pear. & Spear. & Kend. \\
\hline
BLEU1                & -0.078 & -0.167 & -0.125 & -0.113 & -0.157 & -0.118 \\
BLEU3                & 0.221  & 0.074  & 0.025  & 0.135  & 0.041  & 0.014  \\
BLEU4                & 0.189  & 0.063  & 0.012  & 0.106  & 0.035  & 0.008  \\
ROUGE-L                & 0.040   & 0.000      & -0.001 & 0.003  & 0.014  & 0.015  \\
METEOR               & 0.091  & -0.002 & -0.004 & 0.038  & 0.002  & -0.006 \\
BERTScore            & 0.140   & 0.170   & 0.117  & 0.135  & 0.167  & 0.112  \\
BARTScore & & & & & & \\
- Precision            & -0.125 & -0.175 & -0.123 & -0.079 & -0.126 & -0.089 \\
- Recall               & 0.156  & 0.165  & 0.119  & 0.071  & 0.133  & 0.094  \\
- F1                   & -0.081 & -0.145 & -0.105 & -0.058 & -0.124 & -0.084 \\
BARTScore+CNN & & & & & & \\
- Precision            & 0.268  & 0.292  & 0.212  & 0.246  & 0.246  & 0.191  \\
- Recall               & 0.288  & \underline{0.305}  & 0.223  & 0.204  & 0.229  & 0.176  \\
- F1                   & \underline{0.325}  & \textbf{0.339}  & \textbf{0.232}  & 0.243  & \underline{0.256}  & \underline{0.185}  \\
BARTScore+CNN+Para & & & & & & \\
- Precision            & 0.205  & 0.263  & 0.190   & 0.186  & 0.229  & 0.173  \\
- Recall               & 0.273  & 0.282  & 0.184  & 0.182  & 0.212  & 0.149  \\
- F1                   & 0.291  & 0.318  & 0.219  & 0.212  & 0.243  & 0.173  \\
\hline
ChatGPT Multi-Aspect & 0.174  & 0.136  & 0.105  & 0.115  & 0.096  & 0.077  \\
ChatGPT Overall      & 0.196  & 0.101  & 0.086  & 0.13   & 0.075  & 0.067  \\
Vicuna-33b v.1.3 Multi-Aspect  & 0.295  & 0.287  & 0.218  & \underline{0.287}  & \textbf{0.259}  & \textbf{0.215}  \\
Vicuna-33b v.1.3 Overall       & 0.138  & 0.09   & 0.077  & 0.067  & 0.043  & 0.038  \\
GPT-4 Multi-Aspect   & \textbf{0.419}  & 0.274  & \underline{0.228}  & \textbf{0.418}  & 0.204  & 0.178  \\
GPT-4 Overall        & -0.006 & 0.001  & 0.001  & -0.089 & -0.091 & -0.082 \\
Prometheus-13b Multi-Aspect & 0.298 & 0.272 & 0.208 & 0.222 & 0.187 & 0.154 \\
Prometheus-13b Overall & 0.136 & 0.107 & 0.091 & 0.066 & 0.086 & 0.076 \\
\end{tabular}
}
\caption{Correlation of evaluation metric scores to
AMT-generated evaluation scores specifically for ChatGPT-generated candidates; best correlation
is in bold, second is underlined.}
\label{table:fine_grained_chatgpt}
\end{table*}

\begin{table*}[h!]
\ra{1.3}
\huge
\centering
\resizebox{\columnwidth}{!}{%
\begin{tabular}{@{}l l l l l l l @{}}
 \hline
 \multicolumn{7}{c}{Evaluation Metric Correlations (Vicuna v1.3)} \\
 \hline
 \multirow{2}{4cm}{Metric} &\multicolumn{3}{c}{AMT Multi-aspect}  &\multicolumn{3}{c}{AMT Overall} \\
 & Pear. & Spear. & Kend. & Pear. & Spear. & Kend. \\
\hline
BLEU1                & -0.054 & -0.155 & -0.096 & -0.159 & -0.214 & -0.143 \\
BLEU3                & -0.022 & -0.055 & -0.035 & -0.006 & -0.108 & -0.074 \\
BLEU4                & -0.055 & -0.064 & -0.041 & -0.042 & -0.129 & -0.092 \\
ROUGE-L                & -0.036 & -0.135 & -0.104 & -0.147 & -0.247 & -0.166 \\
METEOR               & 0.139  & 0.019  & 0.011  & 0.127  & 0.054  & 0.032  \\
BERTScore            & 0.229  & 0.174  & 0.133  & 0.181  & 0.139  & 0.099  \\
BARTScore & & & & & & \\
- Precision            & -0.218 & -0.170  & -0.104 & -0.328 & -0.298 & -0.211 \\
- Recall               & \textbf{0.442}  & 0.300   & 0.205 & \textbf{0.464}  & \underline{0.356}  & \underline{0.266}  \\
- F1                   & -0.089 & -0.110  & -0.063 & -0.212 & -0.235 & -0.159 \\
BARTScore+CNN & & & & & & \\
- Precision            & 0.291  & 0.219  & 0.158  & 0.215  & 0.145  & 0.118  \\
- Recall               & 0.192  & 0.279  & 0.200    & 0.145  & 0.167  & 0.125  \\
- F1                 & 0.294  & 0.327  & 0.232  & 0.219  & 0.223  & 0.159  \\
BARTScore+CNN+Para & & & & & & \\
- Precision            & 0.159  & 0.202  & 0.135  & 0.147  & 0.163  & 0.127  \\
- Recall               & 0.211  & 0.210   & 0.158  & 0.102  & 0.061  & 0.047  \\
- F1                  & 0.233  & 0.230   & 0.163  & 0.152  & 0.119  & 0.094  \\
\hline
ChatGPT Multi-Aspect & 0.266  & \underline{0.368}  & \underline{0.284}  & 0.214  & 0.236  & 0.182  \\
ChatGPT Overall      & 0.122  & 0.227  & 0.189  & 0.109  & 0.146  & 0.124  \\
Vicuna Multi-Aspect  & \underline{0.409}  & \textbf{0.404}  & \textbf{0.321}  & \underline{0.374}  & \textbf{0.359}  & \textbf{0.293}  \\
Vicuna Overall       & 0.250   & 0.245  & 0.197  & 0.308  & 0.204  & 0.174  \\
GPT-4 Multi-Aspect   & 0.236  & 0.313  & 0.238  & 0.194  & 0.223  & 0.173  \\
GPT-4 Overall        & 0.122  & 0.227  & 0.189  & 0.109  & 0.146  & 0.124 \\
Prometheus-13b Multi-Aspect & 0.172 & 0.120 & 0.093 & 0.104 & 0.018 & 0.004 \\
Prometheus-13b Overall & 0.052 & 0.125 & 0.103 & 0.012 & 0.060 & 0.053 \\
\hline
\end{tabular}
}
\caption{Correlation of evaluation metric scores to
AMT-generated evaluation scores specifically for Vicuna-generated candidates; best correlation
is in bold, second is underlined.}
\label{table:fine_grained_vicuna}
\end{table*}

\setcounter{table}{0}
\renewcommand{\thetable}{H.\arabic{table}}

\begin{table*}[t]
\ra{1.1}
\normalsize
\centering
\resizebox{\linewidth}{!}{%
\begin{tabular}{@{}l  p{40em} @{}}
\hline 
\multicolumn{2}{c}{Multi-aspect Improvement in Prometheus Counter Narrative Evaluation} \\
\hline
 \multirow{2}{4.5cm}{Hate Speech} & \multirow{2}{*}{\parbox{40em}{Jews control the U.S. government.}} \\
 & \\
  \hline
 \multirow{3}{4.5cm}{Generated Counter Narrative} & \multirow{3}{*}{\parbox{40em}{The vast majority of Jews are service members in the military and have worked hard to have a better life for themselves and their families.}} \\
 & \\
 & \\
 \hline
 \multirow{2}{4.5cm}{AMT Overall Score Average} & \multirow{2}{*}{\parbox{40em}{2.33 stars}} \\
 & \\
 \hline
 \multirow{5}{4.5cm}{Prometheus-generated Overall Score and Feedback} & \multirow{5}{*}{\parbox{40em}{5 stars … The counter narrative presents a compelling argument that the majority of Jews are service members in the military, which \textcolor{red}{directly contradicts the hateful claim that Jews control the U.S. government} … \textcolor{red}{demonstrates a deep understanding of the topic and presents strong evidence to support its argument} … The response is clear, concise, and comprehensive, showcasing superior reasoning and a respectful approach …}} \\
 & \\
 & \\
& \\
& \\
\hline
\multicolumn{2}{c}{Prometheus Multi-aspect Scores} \\
\hline
\multirow{4}{4.5cm}{Prometheus-generated Opposition Score and Feedback} & \multirow{4}{*}{\parbox{40em}{1 star … Instead of challenging the notion that Jews control the U.S. government, the response \textcolor{green}{reinforces the idea by suggesting that Jews are primarily service members in the military} …}} \\
& \\
& \\
& \\
\hline
\multirow{4}{4.5cm}{Prometheus-generated Specificity Score and Feedback} & \multirow{4}{*}{\parbox{40em}{Specificity - 1 star … The counter narrative \textcolor{green}{does not provide any in-depth arguments, nuanced reasoning, or supporting evidence} to refute the claim that Jews control the U.S. government.}} \\
& \\
& \\
& \\
\hline

\end{tabular}
}
\caption{Example of improvement in Prometheus evaluation through the use of our multi-aspect framework; Prometheus initially gives a misaligned Overall score by overrating Opposition and Specificity but corrects these ratings when employing our multi-aspect framework.}
\label{table:multi_aspect_prometheus}
\end{table*}

\begin{table*}[t]
\ra{1.1}
\normalsize
\centering
\resizebox{\linewidth}{!}{%
\begin{tabular}{@{}l  p{40em} @{}}
\hline 
\multicolumn{2}{c}{Error - Misunderstanding Counter Narrative} \\
\hline
 \multirow{2}{4.5cm}{Hate Speech} & \multirow{2}{*}{\parbox{40em}{Muslims have no place in civilised Europe.}} \\
 & \\
  \hline
 \multirow{3}{4.5cm}{DialoGPT Generated Counter Narrative} & \multirow{3}{*}{\parbox{40em}{Muslims are the second largest religious group in the European Union, second only to Jews and Hindus.}} \\
 & \\
 & \\
 \hline
 \multirow{3}{4.5cm}{Human Opposition Score and Feedback} & \multirow{3}{*}{\parbox{40em}{1 star. The counter narrative \textcolor{green}{does nothing to refute the claim that Muslims do not belong in Europe, and merely says that they are already there}, which will not be convincing in the least to someone who makes hateful claims.}} \\
 & \\
 & \\
 \hline
 \multirow{3}{4.5cm}{ChatGPT-generated Opposition Score and Feedback} & \multirow{3}{*}{\parbox{40em}{4 stars … It \textcolor{red}{highlights that Muslims are a significant religious group in the European Union, which directly opposes the claim} that they have no place in civilized Europe.}} \\
& \\
& \\
\hline
 \multirow{3}{4.5cm}{GPT4-generated Opposition Score and Feedback} & \multirow{3}{*}{\parbox{40em}{2 stars … The response \textcolor{green}{does not address the sentiment of the hate speech that Muslims do not belong in Europe, it merely states their presence}. }} \\
& \\
& \\
\hline
\end{tabular}
}
\caption{Example of a LLM evaluator providing misaligned evaluation due to misunderstanding the relationship between the counter narrative response and hate speech example; ChatGPT, Vicuna, and Prometheus are more prone to misunderstanding DialoGPT examples in this way.}
\label{table:error_misunderstanding}
\end{table*}

\begin{table*}[t]
\ra{1.1}
\normalsize
\centering
\resizebox{\linewidth}{!}{%
\begin{tabular}{@{}l  p{40em} @{}}
\hline 
\multicolumn{2}{c}{Error - Conflating Aspects} \\
\hline
 \multirow{2}{4.5cm}{Hate Speech} & \multirow{2}{*}{\parbox{40em}{Jews control the U.S. government.}} \\
 & \\
  \hline
 \multirow{5}{4.5cm}{Vicuna Generated Counter Narrative (shortened)} & \multirow{5}{*}{\parbox{40em}{1. Jews are a diverse group of people, just like any other religious or ethnic group, and they don't have any special power to control the U.S. government … 4. It's important to have open and respectful conversations about government policies and actions without making unfounded accusations or spreading hateful stereotypes. }} \\
 & \\
 & \\
 & \\
 & \\
 \hline
 \multirow{3}{4.5cm}{Human Toxicity Score and Feedback} & \multirow{3}{*}{\parbox{40em}{5 stars. The counter makes its arguments in a clear, respectful manner that should not offend anyone.}} \\
 & \\
 & \\
 \hline
 \multirow{3}{4.5cm}{GPT4-generated Toxicity Score and Feedback} & \multirow{3}{*}{\parbox{40em}{4 stars …  \textcolor{red}{Although there is room for improvement in terms of providing more specific examples or evidence}, the overall response is constructive and promotes understanding.}} \\
& \\
& \\
\hline
\end{tabular}
}
\caption{Example of a LLM evaluator providing misaligned evaluation for one aspect due to including information related to another aspect within score; GPT-4 is more prone to this error when evaluating Vicuna-generated candidates.}
\label{table:error_aspects}
\end{table*}

\begin{table*}[t]
\ra{1.1}
\normalsize
\centering
\resizebox{\linewidth}{!}{%
\begin{tabular}{@{}l  p{40em} @{}}
\hline 
\multicolumn{2}{c}{GPT-4 Error - Underrating Vicuna} \\
\hline
 \multirow{2}{4.5cm}{Hate Speech} & \multirow{2}{*}{\parbox{40em}{Islam is not a peaceful religion, it is full of hate and violence.}} \\
 & \\
  \hline
 \multirow{5}{4.5cm}{Vicuna Generated Counter Narrative (shortened)} & \multirow{5}{*}{\parbox{40em}{1. Islam is a religion of peace, and it is important to recognize that the actions of a few do not represent the entire faith. 
2. The Quran teaches peace, compassion, and mercy, and there are many verses promoting these values. 
3. Many Muslims around the world are working to promote peace, justice, and understanding among people of all faiths and backgrounds …
}} \\
 & \\
 & \\
 & \\
 & \\
 \hline
 \multirow{4}{4.5cm}{Human Specificity Score and Feedback} & \multirow{4}{*}{\parbox{40em}{5 stars. The hate speech is about how Islam is hateful and violent. The counter goes over several points to refute that, from what the religion itself actually teaches to the large amounts of peaceful, nonviolent things its adherents promote in reality.}} \\
 & \\
 & \\
 & \\
 \hline
 \multirow{3}{4.5cm}{GPT4-generated Toxicity Score and Feedback} & \multirow{3}{*}{\parbox{40em}{2 stars. The counter narrative response provides some degree of specificity … \textcolor{red}{However, the arguments presented are relatively broad and lack depth} … 
}} \\
& \\
& \\
\hline
\end{tabular}
}
\caption{Example of a GPT-4 providing misaligned evaluation due to underrating Vicuna-generated candiates relative to AMT annotation.}
\label{table:gpt4_error}
\end{table*}

\setcounter{table}{0}
\renewcommand{\thetable}{I.\arabic{table}}

\begin{table*}[h!]
\ra{1.2}
\large
\centering
\resizebox{\linewidth}{!}{%
\begin{tabular}{@{}l l l l l l l l @{}}
\hline
 \multicolumn{8}{c}{Average AMT Scores} \\
 \hline
Generation Model & Evaluation Approach & Opposition &  Relatedness & Specificity & Toxicity & Fluency & Overall \\
\hline
DialoGPT        &                 &      &      &      &      &      &      \\
                & Human        & 2.76  & 3.22 & 1.88  & 3.58 & 3.81 & 2.04 \\
                & LLM Evaluators & & & & & & \\
                & - GPT-4           & 2.35 (-0.41) & 2.88 (-0.34)  & 1.68 (-0.20) & 4.33 (+0.75) & 2.88 (-0.93) & 1.82 (-0.22) \\
                & - ChatGPT         & 3.18 (+0.42) & 3.50 (+0.28)  & 2.35 (+0.47) & 3.38 (-0.20) & 2.92 (-0.89) & 2.47 (+0.43)  \\
                & - Vicuna-33b v1.3 & 2.40 (-0.36)  & 2.47 (-0.75) & 1.58 (-0.30) & 3.48 (-0.10) & 3.15 (-0.66) & 1.42 (-0.62)  \\
                & - Prometheus-13b & 1.43 (-1.33) & 1.83 (-1.39) & 1.55 (-0.33)  & 3.53 (-0.05) & 3.07 (-0.74)  & 2.45 (+0.41)  \\
ChatGPT         &                 &      &      &      &      &      &      \\
                & Human           & 4.78 & 4.71 & 4.18 & 4.64 & 4.77 & 4.36 \\
                & LLM Evaluators & & & & & & \\
                & - GPT-4           & 4.95 (+0.17) & 4.95 (+0.24) & 3.70 (-0.48)  & 5.00 (+0.36)    & 5.00 (+0.23)   & 4.85 (+0.49) \\
                & - ChatGPT         & 4.02 (-0.76) & 4.13 (-0.58) & 3.42 (-0.76) & 4.15 (-0.49) & 4.02 (-0.75) & 3.88 (-0.48) \\
                & - Vicuna-33b v1.3 & 5.00 (+0.22)    & 4.78 (+0.07) & 3.95 (-0.23) & 5.00 (+0.36)    & 5.00 (+0.23)  & 4.63 (+0.27) \\
                & - Prometheus-13b & 4.20 (-0.58) & 4.92 (+0.21) & 4.03 (-0.15) & 4.97 (-0.33) & 4.33 (-0.44) & 4.82 (-0.46) \\
Vicuna-33b v1.3 &                 &      &      &      &      &      &      \\
                & Human           & 4.44 & 4.54 & 3.98 & 4.86 & 4.34 & 4.02 \\
                & LLM Evaluators & & & & & & \\
                & - GPT-4           & 3.90 (-0.54)  & 4.03 (-0.51) & 3.13 (-0.85) & 4.05 (-0.81) & 3.72 (-0.62) & 3.55 (-0.47) \\
                & - ChatGPT         & 3.92 (-0.52) & 4.05 (-0.49) & 3.13 (-0.85) & 4.05 (-0.81) & 3.70 (-0.64) & 3.57 (-0.45) \\
                & - Vicuna-33b v1.3 & 4.95 (+0.51) & 4.48 (-0.06) & 3.32 (-0.66) & 4.72 (-0.14) & 4.60 (+0.26)  & 3.92 (-0.10) \\
                & - Prometheus-13b & 4.05 (-0.39) & 5.00 (-0.46) & 3.95 (-0.03) & 5.00 (-0.14) & 4.33 (-0.01) & 4.77 (-0.75) \\
\hline
\end{tabular}
}
\caption{Average score given to counter narratives generated by each generation model used in our evaluation set including average scores given from each LLM evaluator.}
\label{table:average_score_llm}
\end{table*}

\begin{table*}[h!]
\ra{1.2}
\centering
\resizebox{\linewidth}{!}{%
\begin{tabular}{@{} lllllllll @{}}
\hline
 \multicolumn{9}{c}{Mean Absolute Error} \\
\hline
Generation Model & Evaluation Approach & Opposition & Relatedness & Specificity & Toxicity & Fluency & Aspect Average & Overall \\
\hline
DialoGPT        &                 &      &      &      &      &      &      &      \\
                & GPT-4           & \textbf{0.77} & \textbf{1.01} & \textbf{0.54} & 0.91 & \textbf{1.15} & \textbf{0.52} & \textbf{0.53} \\
                & ChatGPT         & 1.02 & 1.03 & 0.9  & 0.91 & 1.26 & 0.66 & 0.87 \\
                & Vicuna-33b v1.3 & 1.01 & 1.2  & 0.79 & \textbf{0.83} & \textbf{1.15} & 0.74 & 0.95 \\
                & Prometheus-13b & 1.48 & 2.18 & 0.97 & 1.07 & 1.36 & 1.09 & 1.33 \\
ChatGPT         &                 &      &      &      &      &      &      &      \\
                & GPT-4           & \textbf{0.21} & 0.29 & \textbf{0.67} & \textbf{0.35} & \textbf{0.23} & \textbf{0.22} & 0.66 \\
                & ChatGPT         & 0.81 & 0.73 & 0.9  & 0.69 & 0.75 & 0.7  & 0.64 \\
                & Vicuna-33b v1.3 & 0.22 & 0.39 & 0.7  & 0.36 & \textbf{0.23} & 0.25 & \textbf{0.61} \\
                & Prometheus-13b & 0.68 & \textbf{0.25} & 0.69 & 0.37 & 0.57 & 0.32 & 0.62 \\
Vicuna-33b v1.3 &                 &      &      &      &      &      &      &      \\
                & GPT-4           & 0.75 & 0.71 & 1.2  & 0.92 & 0.89 & 0.73 & 0.77 \\
                & ChatGPT         & 0.74 & 0.69 & 1.19 & 0.92 & 0.89 & 0.73 & \textbf{0.76} \\
                & Vicuna-33b v1.3 & \textbf{0.57} & 0.59 & \textbf{0.99} & 0.38 & \textbf{0.44} & \textbf{0.3}  & 0.82 \\
                & Prometheus-13b & 0.84 & \textbf{0.46} & \textbf{0.99} & \textbf{0.14} & 0.49 & 0.41 & 0.91 \\
All Models      &                 &      &      &      &      &      &      &      \\
                & GPT-4           & \textbf{0.58} & \textbf{0.67} & \textbf{0.81} & 0.73 & 0.76 & 0.49 & \textbf{0.65} \\
                & ChatGPT         & 0.86 & 0.82 & 1    & 0.84 & 0.97 & 0.69 & 0.76 \\
                & Vicuna-33b v1.3 & 0.6  & 0.73 & 0.83 & \textbf{0.52} & \textbf{0.61} & \textbf{0.43} & 0.79 \\
                & Prometheus-13b & 1 & 0.96 & 0.89 & 0.53 & 0.81 & 0.61 & 0.95 \\
\hline
\end{tabular}
}
\caption{Mean absolute error for scores generated by each LLM evaluator in our study per generation approach as well as for all candidates generated.}
\label{table:mae}
\end{table*}

\begin{table*}[h!]
\ra{1.2}
\large
\centering
\resizebox{\linewidth}{!}{%
\begin{tabular}{@{}l l l l l l l  l@{}}
\hline
 \multicolumn{8}{c}{Average AMT Scores} \\
 \hline
Generation Model & Opposition & Relatedness & Specificity & Toxicity & Fluency & Aspect Average & Overall \\
\hline
DialoGPT         & 2.76 ± 1.33   & 3.22 ± 1.04      & 1.88 ± 0.76     & 3.58 ± 1.20    & 3.81 ± 1.02  & 3.05 ± 0.73  & 2.04 ± 0.83   \\
ChatGPT          & \textbf{4.78 ± 0.35}      & \textbf{4.71 ± 0.54}       & \textbf{4.18 ± 0.72}       & 4.64 ± 0.47  & \textbf{4.77 ± 0.29}   & \textbf{4.62 ± 0.32} & \textbf{4.36 ± 0.60}   \\
Vicuna-33b v1.3  & 4.44 ± 0.60     & 4.54 ± 0.64    & 3.98 ± 0.86    & \textbf{4.86 ± 0.36}    & 4.34 ± 0.75  & 4.43 ± 0.43  & 4.02 ± 0.71 \\
All Models  & 3.99 ± 1.24 & 4.16 ± 1.02 & 3.34 ± 1.3 & 4.36 ± 0.96	 & 4.31	±  0.85 & 4.03 ± 0.87 & 3.47 ± 1.25 \\
\hline
\end{tabular}
}
\caption{Average score given from AMT workers to counter narratives generated by each generation model used in our evaluation set including standard deviation.}
\label{table:standard_deviation}
\end{table*}

\end{document}